\documentclass[letterpaper]{article} % DO NOT CHANGE THIS
\usepackage{aaai25}  % DO NOT CHANGE THIS
\usepackage{times}  % DO NOT CHANGE THIS
\usepackage{helvet}  % DO NOT CHANGE THIS
\usepackage{courier}  % DO NOT CHANGE THIS
\usepackage[hyphens]{url}  % DO NOT CHANGE THIS
\usepackage{graphicx} % DO NOT CHANGE THIS
\usepackage{natbib}  % DO NOT CHANGE THIS AND DO NOT ADD ANY OPTIONS TO IT
\usepackage{caption} % DO NOT CHANGE THIS AND DO NOT ADD ANY OPTIONS TO IT
\usepackage{algorithm}
\usepackage{algorithmic}
\usepackage{multirow}
\usepackage{xcolor}
\usepackage{color, colortbl}
\usepackage{todonotes}
\usepackage{amsmath}
\usepackage{comment}
\usepackage{booktabs}

\usepackage{newfloat}
\usepackage{listings}
\DeclareCaptionStyle{ruled}{labelfont=normalfont,labelsep=colon,strut=off} % DO NOT CHANGE THIS
\floatstyle{ruled}
\newfloat{listing}{tb}{lst}{}
\floatname{listing}{Listing}
\title{Multilingual LLMs Inherently Reward In-Language Time–Sensitive Semantic Alignment for Low-Resource Languages}
\author {
    Ashutosh Bajpai\textsuperscript{\rm 1,2},
    Tanmoy Chakraborty\textsuperscript{\rm 1}
}
\affiliations {
    \textsuperscript{\rm 1} Indian Institute of Technology Delhi, India\\
    \textsuperscript{\rm 2} Wipro Research, India\\
    eez228482@ee.iitd.ac.in, tanchak@ee.iitd.ac.in
}

\newcommand{\llms}
{LLMs}
\newcommand{\tempreason}
{TEMPREASON}
\newcommand{\mtempreason}
{\texttt{mTEMPREASON}}
\newcommand{\newmodel}
{\texttt{CLiTSSA}}
\newcommand{\xinsta}
{X-InSTA}
\newcommand{\xicl}
{X-ICL}

\newcommand{\newmodelfull}{\textbf{C}ross-\textbf{L}ingual \textbf{T}ime-\textbf{S}ensitive \textbf{S}emantic \textbf{A}lignment}

\definecolor{LightGreen}{rgb}{0.6,9,0.6}
\definecolor{LighterGreen}{rgb}{0.85,9,0.85}
\definecolor{Green}{rgb}{0.2,9.5,0.3}
\definecolor{LightGray}{rgb}{0.9,0.9,0.9}
\definecolor{LightRed}{rgb}{0.99,0.8,0.8}
\usepackage{bibentry}
\begin{document}

\maketitle

\begin{abstract}
The unwavering disparity in labeled resources between resource-rich languages and those considered low-resource remains a significant impediment for Large Language Models (LLMs). Recent strides in cross-lingual in-context learning (X-ICL), mainly through semantically aligned examples retrieved from multilingual pre-trained transformers, have shown promise in mitigating this issue. However, our investigation reveals that LLMs intrinsically reward in-language semantically aligned cross-lingual instances over direct cross-lingual semantic alignments, with a pronounced disparity in handling time–sensitive queries in the X-ICL setup. Such queries demand sound temporal reasoning ability from LLMs, yet the advancements have predominantly focused on English. This study aims to bridge this gap by improving temporal reasoning capabilities in low-resource languages. To this end, we introduce \mtempreason, a temporal reasoning dataset aimed at the varied degrees of low-resource languages and propose \newmodelfull\ (\newmodel), a novel method to improve temporal reasoning in these contexts. To facilitate this, we construct an extension of \mtempreason\ comprising pairs of parallel cross–language temporal queries along with their anticipated in-language semantic similarity scores. Our empirical evidence underscores the superior performance of \newmodel\ compared to established baselines across three languages -- Romanian, German, and French,  encompassing three temporal tasks and including a diverse set of four contemporaneous LLMs. This marks a significant step forward in addressing resource disparity in the context of temporal reasoning across languages.
\end{abstract}

% Uncomment the following to link to your code, datasets, an extended version or similar.
%
% \begin{links}
%     \link{Code}{https://aaai.org/example/code}
%     \link{Datasets}{https://aaai.org/example/datasets}
%     \link{Extended version}{https://aaai.org/example/extended-version}
% \end{links}

\section{Introduction}
\label{section:intro}
In the evolving landscape of Large Language Models (\llms), temporal reasoning requires models to comprehend and interpret the significant subtleties inherent in time–time, time–event and event–event correlations \cite{chen2021datasetansweringtimesensitivequestions,dhingra-etal-2022-time}. Temporality is a crucial dimension of information that evolves through creation, maintenance, and obsolescence. Enhancing LLMs with this faculty augments their analytical capabilities, paving the way for addressing intricate challenges prevalent in domains sensitive to temporal dynamics, such as finance, healthcare, legal studies, and archaeology. Furthermore, addressing low-resource languages in LLMs is crucial for computational linguistics, given their paucity of data and digital infrastructure \cite{cahyawijaya2023nusacrowdopensourceinitiative,asai2023buffetbenchmarkinglargelanguage,adilazuarda2024lingualchemyfusingtypologicalgeographical}. Enhancing LLMs for these languages improves not just genuine linguistic inclusivity but also their application and acceptance across diverse cultural landscapes. The discourse on enhancing temporal reasoning in LLMs has, until now, been predominantly focused on English. Our work seeks to alleviate this disparity by propelling temporal reasoning in low-resource languages.

\subsubsection{Cross-Lingual In-Context Prompting.} Recent advancements in in-context learning (ICL), prompted by the advent of LLMs, have shown promising results  \cite{zhao2021calibrateuseimprovingfewshot,lin2022fewshotlearningmultilinguallanguage,liu-etal-2022-makes,zhang2022differentiablepromptmakespretrained}. 
The stark contrast in annotated data availability among languages accentuates the usage of high-resource linguistic contexts for addressing tasks in low-resource languages. The ICL approach was adapted by \citet{winata-etal-2021-language} for cross-lingual (X-ICL) applications by randomly selecting examples from a resource-rich language to support queries in a language with limited resources.

\begin{figure*}[t]
\centering
\includegraphics[width=1\textwidth]{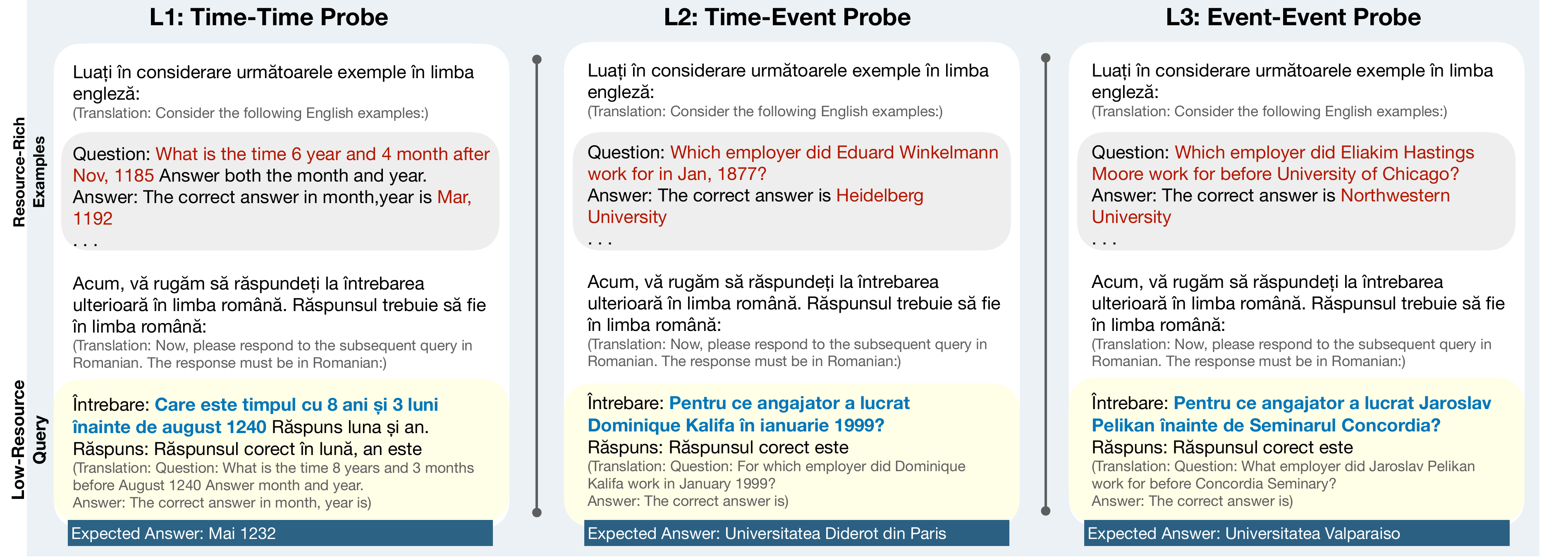} % Reduce the figure size so that it is slightly narrower than the column.
\caption{A working example of the low-resource cross-lingual prompting across three temporal tasks: L1, L2, and L3 in Romanian. The translations included in small brackets are not integral to the prompt; their purpose is solely to enhance readability.} 
\label{fig:main}
\vspace{-5mm}
\end{figure*}

Figure \ref{fig:main} shows three levels of temporal queries - L1 (Time–Time), L2 (Time–Event), and L3 (Event–Event) and the expected model responses: time information for L1 queries, an event for L2 given time, and for L3, an event in response to another event, without explicit temporal details in either input or output. An illustrative L1 query in English (\textit{``What is the time 6 years and 4 months after Nov, 1185''}) with an associated answer (\textit{``Mar, 1192''}) serves as additional context for a corresponding L1 query in Romanian (\textit{``Care este timpul cu 8 ani și 3 luni înainte de august 1240''}). Subsequent research suggested that cross-lingual examples that are semantically aligned can significantly enhance performance compared to the arbitrarily selected examples \cite{tanwar2023multilingualllmsbettercrosslingual}. Further development also indicates that semantic similarity alone doesn't ensure the optimal performance, stressing the necessity for a learning-based retrieval model. \cite{lin2024xamplerlearningretrievecrosslingual}.

\subsubsection{Challenges in Cross-Lingual Semantic Alignment.} Cross-lingual in-context approaches rely on the contextual semantic profoundness embedded within multilingual pre-trained encoder-only transformers, reflected through their embedding space, for retrieving semantically akin examples. Nonetheless, the disparity in linguistic distribution within the pre-training dataset, favoring resource-rich languages over those with fewer resources, significantly hinders efficient cross-lingual semantic alignment within their embedding space%as highlighted by \citep{cahyawijaya2024llmsfewshotincontextlowresource}%
, especially concerning time–sensitive queries. The ensuing analysis is conducted to validate the aforementioned hypothesis in a temporal context.  

The objective is to assess the efficacy of multilingual sentence-BERT \cite{reimers2019sentencebertsentenceembeddingsusing} in retrieving semantically akin cross-lingual in-context examples for L1 task, considering Romanian, German, and French as low- and English as a high-resource language. %utilizing cosine similarity. 
The investigation juxtaposes two distinct approaches: firstly, \textit{cross-lingual-similarity}, wherein the top-3 English instances are directly retrieved for a low-resource query by ranking the similarity scores between a low-resource query and English instances. Secondly, \textit{in-language-similarity}, where this is achieved by initially translating the English example dataset into low-resource languages. Next, the top-3 instances are sourced using similarity scores between a low-resource query and the translated English examples dataset. Subsequently, the translated examples are replaced with their corresponding English instances, thus retrieving the English examples with the highest in-language similarity. Performances are quantified using F1 scores and exact match (EM) scores. The results presented in Table \ref{tab:hypothesis} indicate that multilingual encoder-only transformers exhibit profound semantic similarity for temporal queries in an in-language similarity framework, outperforming their counterparts in a cross-lingual similarity context when identifying semantically akin examples for the cross-lingual temporal reasoning task across languages%, especially in case of the explicit temporal tasks -- L1 and L2
. Consequently, this underscores the need to evaluate and enhance the cross-lingual, time–sensitive semantic context of retrieval models. Nonetheless, the availability of data for such alignment in low-density languages is limited. 
%@{\extracolsep{8pt}}

\begin{table}[t]
\centering
\small
%\resizebox{.95\columnwidth}{!}{
\begin{tabular}%{l@{\hspace{0.5cm}}|l@{\hspace{1cm}}|c|c}
{l@{\hspace{0.3cm}}|l@{\hspace{0.5cm}}|c|c}
\toprule
    \textbf{Task} & \textbf{Settings} & \textbf{F1.} & \textbf{EM} \\
    \hline
\multirow{2}*{{Romanian}}& Cross-Lingual-Similarity & 33.65	&10.15\\
& In-Language-Similarity & \textbf{43.48}	& \textbf{18.65}\\
	\midrule
\multirow{2}*{{German}}& Cross-Lingual-Similarity &56.63&	35.45\\
& In-Language-Similarity & \textbf{64.77}	& \textbf{46.35}\\
	\midrule
\multirow{2}*{{French}}& Cross-Lingual-Similarity &46.62&	22.05\\
&  In-Language-Similarity& \textbf{60.50}	&\textbf{37.17}\\

\bottomrule
\end{tabular}
\caption{Analyzing the impact of in-language versus cross-language similarities in retrieving semantically similar akin examples in a three-shot cross-lingual setup for the L1 task across languages using LLaMA3-8B}
\label{tab:hypothesis}
\vspace{-5mm}
\end{table}

\subsubsection{Our Proposed Method.}

We start with the development of first-of-its-kind a comprehensive benchmark dataset, \mtempreason, to evaluate temporal reasoning for limited resource languages -- Romanian, German, and French across a diverse set of LLMs. Further, we endeavor to devise an efficacious novel cross-lingual retriever, \newmodel\ (\newmodelfull), for handling time–sensitive queries from low-resource languages, addressing the aforementioned challenges in a cross-lingual context. To achieve this, drawing inspiration from \citet{yamada2024leiafacilitatingcrosslingualknowledge}, we elect to apply the transfer of profound semantic space knowledge within a language to a cross-lingual semantic space for queries influenced by temporality. Consequently, we adopt a supervised fine-tuning approach that necessitates an additional dataset to facilitate this transition. To this end, we curate an extension of the \mtempreason\ dataset comprising parallel sentences for Romanian-English, German-English, and French-English pairs, accompanied by their anticipated similarity scores in the semantic space of the low-resource language. By employing this curated dataset, the transition of the semantic context from a monolingual to a cross-lingual embedding sphere is attained.

Remarkably, for temporal queries, \newmodel\ outperforms the arbitrary cross-lingual in-context benchmark, demonstrating relative mean F1 score enhancements of $11.41\%$, $30.77\%$, and $62.92\%$ for Romanian, German, and French, respectively. Additionally, it evidences a significant improvement in relative mean F1 score of $6.38\%$, $5.98\%$, and $20.93\%$ compared to the contemporary cross-lingual in-context baselines for Romanian, German, and French.

\begin{table}[t]
\centering
\small
%\resizebox{.95\columnwidth}{!}{
\begin{tabular}{l@{\hspace{0.7cm}}|@{\hspace{0.4cm}}c|@{\hspace{0.4cm}}c|@{\hspace{0.4cm}}c}
\toprule
   & \textbf{Train} & \textbf{Dev} & \textbf{Test} \\
    \hline
Time Range &1014-2022 &634-2023& 998-2023\\
L1& 400,000& 4,000& 4,000\\
L2&16,017 &5,521 &5,397\\
L3&13,014 &4,437 &4,426\\

\bottomrule
\end{tabular}
\caption{Dataset statistics for \mtempreason.}
\label{tab:datastas}
\vspace{-3mm}
\end{table}

Our contributions are summarized below\footnote{
Source code and dataset are available at \url{https://github.com/ab-iitd/clitssa}.}--

\begin{itemize}
    \item We develop a dataset centered around temporal reasoning, \mtempreason, elusively designed for varying degrees of limited resource languages.
    \item Our findings reveal that multilingual transformers exhibit superior in-language semantic similarity over cross-lingual similarity context for temporal queries, especially explicit ones, in the X-ICL setup.
    \item We introduce \newmodel\ to enhance temporal reasoning capabilities within \llms\ for low-resource languages. Consequently, we develop an extension of \mtempreason\ comprised of paired cross-lingual time–sensitive queries with corresponding similarity scores. Our empirical analysis demonstrates that \newmodel\ significantly outperforms the contemporary baselines.
\end{itemize}

\section{Benchmark For Low-Resource Temporal Reasoning}
\label{section:resource}

In our study of temporal reasoning, the \tempreason\ \cite{tan2023benchmarkingimprovingtemporalreasoning} stands out as a recent, comprehensive resource, providing multifaceted temporality across an extended time frame. Therefore, we select it to develop the first multilingual, low-resource dataset for temporal reasoning. To this end, we employ the T5 model \cite{raffel2023exploringlimitstransferlearning} to automatically translate the dataset from English language to Romanian, German and French languages.
\subsection{The \mtempreason\ Dataset}
\tempreason\ encompasses tasks categorized into three levels of temporal complexity, namely time–time, time–event, and event–event relationships, corresponding to Levels L1, L2, and L3, respectively. The dataset's statistical information, along with the partition into training (Train), development (Dev), and test set (Test), is detailed in Table \ref{tab:datastas}. Employing a concise prompt prefix, ``Translate the following sentences from English to the \{Target Language\},'' we employ the T5 model to translate this dataset into a selection of languages with varying degrees of limited resources, specifically French, German, and Romanian. We opt for these three languages in this work as they provide varied levels of limited resources compared to English. Romanian, German, and French have $98.3\%$, $89.5\%$, and $78\%$ fewer speakers compared to English\footnote{https://www.ethnologue.com/insights/ethnologue200}, respectively. 

\begin{table}[t]
\centering
\small
%\resizebox{.95\columnwidth}{!}{
\begin{tabular}{l|l|c|c|c|c}
\toprule
  
    \textbf{Metric} & \textbf{Setting} & \textbf{Ro.} & \textbf{Ge.} & \textbf{Fr.}& \textbf{Avg.}\\
    \hline
TSR& Auto &99.00 &97.57&97.79&$98.11\pm1.8$\\

    \midrule
{BTA}& BLEU-3 & 51.12&56.47&43.63&50.41\\

\bottomrule
\end{tabular}
\caption{Quality assessment of \mtempreason's translations: employing automated verification for \textit{Translation Success Rate (TSR in \%)}, and applying BLEU-3 and manual review standards for \textit{Back-Translation Accuracy (BTA)} evaluation across languages— Romanian (Ro.), German (Ge.) and French(Fr.), averaging over temporal tasks.} 
\label{tab:quality}
\vspace{-3mm}
\end{table}

\subsection{Data Quality}

The \mtempreason\ dataset was constructed by a linguist who specialized in NLP\footnote{A male expert, within the 30-40 age bracket.}. We employed back-translation-based \citep{Miyabe2015EvaluationOT} evaluation to ensure the superior quality of the proposed dataset. A random selection of $100$ query examples was made from  \mtempreason\ for each of the translated languages -— Romanian, German, and French, across temporal tasks within the test dataset. These queries underwent back-translation\footnote{https://translate.google.com} into the source language (English) and were subsequently compared to their original counterparts in the resource-rich language (English) to assess fidelity and coherence. The analysis employed the BLEU-3 metric for quantitative evaluation. In addition, successful translation from the source to the target languages was noted, regardless of translation quality. The translation success rate was documented using an automated approach by employing a language detection library\footnote{https://pypi.org/project/langdetect} across the entire test dataset. In Table \ref{tab:quality}, the mean automated translation success rate was recorded at $98.11\pm1.83\%$. Concurrently, a mean BLEU-3 score of $50.41$ was observed for assessing back-translation-based accuracy.

\subsection{Problem Setting}

The Time–Sensitive Question Answering (TSQA) task requires that the LLMs generate an accurate answer in response to a temporal query. This answer may be a temporal delineation or an event, depending on the structure of the query, which can include time–time, time–event, and event–event scenarios. Our experiments are conducted in a closed-book environment, requiring LLMs to deliver precise facts without reliance on external contexts. As illustrated in Figure \ref{fig:main}, to demonstrate the prompt in a cross-lingual in-context learning framework, a few-shot example from a resource-rich language, accompanied by the query in a low-resource language, is provided. 

In this study, we consider the following baselines-
\begin{itemize}
    \item \textbf{Cross-Lingual In-Context Learning (\xicl) \cite{winata-etal-2021-language}.} The model is primed with limited examples from resource-rich language serving as demonstrations, along with a query in low-resource language.
    \item \textbf{\xinsta\ (Semantic Aligner) \cite{tanwar2023multilingualllmsbettercrosslingual}.} \xinsta\ has advanced the \xicl\ method by introducing a retrieval of semantically akin examples for queries across languages, leveraging label space alignment.
\end{itemize}

\section{Method}
\label{section:method}

\subsection{Primer}
Let us consider a resource-rich source dataset $D_{r}$, containing pairs of queries and answers ($q_i^r$,$a_i^r$) for each $i \in m$, with $m$ representing the total number of samples within $D_{r}$. Additionally, let $D_{l}$ be a low-resource language dataset, which similarly comprises query and answer pairs ($q_j^l$,$a_j^l$) for each $j \in n$, where $n$ signifies the total sample count in $D_{l}$. Within a conventional ICL framework, $K$ arbitrary instances of question-answer pairs are selected from $D_{r}$, designated as context $C$ for a low-resource query $q_x^l$ from $D_{l}$, with $x \in n$. The goal is to optimize the expected value of $a_x^l$, given context $C$ and the query input as illustrated in Equation \ref{eq:maxlikelihood}, where $A^l$ represents the vocabulary space corresponding to query $q_x^l \in D_{l}$.
\begin{equation}
   a_x^l  = \arg\max_{a^l \in A^l} p(a^l |C, q_x^l)
   \label{eq:maxlikelihood}
\end{equation}
In the case of a semantic aligner, $C$ is constructed to maximize the semantic alignment between query $q_x^l$ and context $C$. Hence, we introduce $e_{q_x^l}$ as a dense embedding representation produced by a multilingual pre-trained transformer for query $q_x^l$. Correspondingly, for each $i\in m$, $e_{q_i^r}$ represents the embeddings for queries within a resource-rich dataset $D_{r}$. Furthermore, $f(s)$ and $f(d)$ denote similarity and distance functions, respectively, such that for cosine similarity, $f(s) = 1 - f(d)$; a lesser distance implies greater similarity. The overarching goal is to identify a set of $K$ examples, denoted as $S_K$, where the semantic similarity surpasses that of other dataset examples, as shown in Equation \ref{eq:semanticsim}. 
\begin{equation}
\begin{split}
   S_K = \{(q_k^r, &a_k^r) \forall k \in \{1,\dots, K\} , \\ &\text{if} \, f(s)_{q_k^r,q_x^l} \geq f(s)_{q_z^r,q_x^l} \forall z \in m \textrm{ and } z \notin K\}
   \end{split}
   \label{eq:semanticsim}
\end{equation}
The variable $f(s)_{q_k^r,q_x^l}$ denotes the semantic similarity between a low-resource query $q_x^l$ and an example query $q_k^r$ from a resource-rich language. To calculate $f(s)_{q_k^r,q_x^l}$, the procedure stats with the extraction of dense embeddings, $e_{q_k^r}$ and $e_{q_x^l}$ for the input queries $q_k^r$ and $q_x^l$, respectively. This extraction is performed utilizing a multilingual pre-trained transformer model. Subsequently, the function $f(d)$ is applied to these embeddings to measure the distance, which ultimately yields the value of $f(s)$.

\subsection{\newmodelfull\ (\newmodel)}

\subsubsection{Objective.}
We introduce \newmodel\ to augment the semantic similarity context of time–sensitive queries within the cross-lingual embedding space. We elect to employ the transfer of the comprehensive time–sensitive semantic knowledge from an in-language embedding space to a cross-lingual embedding space. For this purpose, we have embraced a supervised fine-tuning approach, making use of a training dataset comprised of sentence pairs along with their associated labeled scores, quantifying the expected time–sensitive semantic similarity among pairs of queries. The objective is to achieve an effective cross-lingual time–sensitive contextual alignment of temporal queries for \llms, thus boosting in-context performance.

\begin{table*}[t!]
\centering
\small
%\resizebox{.95\columnwidth}{!}{
\begin{tabular}{l@{\hspace{0.4cm}}|l@{\hspace{0.4cm}}|ccc l@{\hspace{0.4cm}}|@{\hspace{0.4cm}}cccc}
\toprule
&  & \multicolumn{4}{c@{\hspace{0.4cm}}|@{\hspace{0.4cm}}}{\textbf{F1.}} & \multicolumn{4}{c}{\textbf{EM}} \\
\cline{3-10}

\textbf{Task}& \textbf{Method} &\textbf{French} &\textbf{German}& \textbf{Romanian}& \textbf{Avg.}&  \textbf{French}& \textbf{German}& \textbf{Romanian}& \textbf{Avg.}\\

\hline

\multirow{4}*{L1} &\xicl &33.60&	45.33	&34.17&	37.70&	14.85	&22.79&	08.80	&15.48\\
							
 & \xinsta$^\uparrow$ &46.62	&56.63&	33.65	&45.63	&22.05	&35.45	&10.15&	22.55\\

 %& \xampler &33.71 & & & 3.57& & \\

 & $\newmodel$ &\textbf{57.15}&	59.77&	37.16&	51.36&	\textbf{32.57}&	39.3&	13.45&	28.44\\

 & $\newmodel^*$ & 55.04	&\textbf{63.50}&	\textbf{37.78}&	\textbf{52.11}&	31.62	&\textbf{45.15}&	\textbf{13.70}&\textbf{30.16}\\

%& & & & & & & &\\
\midrule
%\cline{2-10}
\multirow{4}*{L2} & \xicl  &11.00	&08.96&	10.99&	10.32&	03.57&	03.73&	03.97&	03.76\\
		
 & \xinsta$^\uparrow$ &11.92&	12.45&	11.40&	11.92&	04.55&	05.18&	03.83	&04.52 \\

 %& \xampler &11.92 & & &4.55 & & \\

 & \newmodel &
\textbf{15.23}&\textbf{	14.02}&	11.42	&\textbf{13.56}&\textbf{05.81}	&\textbf{05.24}&	03.87&	04.97\\

 & $\newmodel^*$ &  15.04	&13.53&	{\textbf{11.71}}	& 13.43&	05.74	&05.20&	{\textbf{04.03}}&{\textbf{04.99}}\\

%& & & & & & & &\\
\midrule
%\cline{2-10}
 \multirow{4}*{L3} & \xicl  & 17.07	&16.18	&18.84&17.36	&07.28&	05.77	&08.82	&07.29\\

 & \xinsta$^\uparrow$ &17.74&	18.35&	22.18&	19.42	&10.55	&\textbf{09.08}&	12.85&	10.83\\

 %& \xampler &17.74 & & &10.55 & & \\

 & \newmodel &19.87&	18.52&	\textbf{22.94}&	\textbf{20.44}&	11.22	&08.85	&\textbf{13.33}&	\textbf{11.13}\\
	
& $\newmodel^*$ &  \textbf{19.92}&	{\textbf{18.69}}&	22.53&20.38&		\textbf{11.45}&	08.72	&13.23&11.13\\

\midrule

&$\overline{\Delta}_{\newmodel - \uparrow}$& \textcolor{blue}{${5.32\uparrow}$} &\textcolor{blue}{${1.62\uparrow}$}&\textcolor{blue}{${1.43\uparrow}$}&	\textcolor{blue}{${2.79\uparrow}$}&	\textcolor{blue}{${4.15\uparrow}$}&\textcolor{blue}{${1.23\uparrow}$}&\textcolor{blue}{${1.27\uparrow}$}&	\textcolor{blue}{${2.22\uparrow}$}\\

&$\overline{\Delta}_{\newmodel^{max} - \uparrow}$& \textcolor{blue}{$\boldsymbol{5.34\uparrow}$} &\textcolor{blue}{$\boldsymbol{2.92\uparrow}$}&\textcolor{blue}{$\boldsymbol{1.73\uparrow}$}&	\textcolor{blue}{$\boldsymbol{3.04\uparrow}$}&	\textcolor{blue}{$\boldsymbol{4.22\uparrow}$}&\textcolor{blue}{$\boldsymbol{3.17\uparrow}$}&\textcolor{blue}{$\boldsymbol{1.41\uparrow}$}&	\textcolor{blue}{$\boldsymbol{2.79\uparrow}$}\\

\bottomrule
%\hline
\end{tabular}
\caption{Comparison of F1 and EM (Exact Match) scores across different prompting strategies for temporal tasks and languages in a three-shot setup employing LLaMA3-8B. The strategies include \textit{X-ICL} and \textit{X-InSTA}, representing random and semantically aligned cross-lingual baselines, respectively, while $\newmodel^*$ denotes an integrated retriever trained across languages and tasks, and $\newmodel$ indicates a language and task-specific retriever. $\overline{\Delta}$ represents mean improvement for languages across temporal tasks and $\newmodel^{max}$ representing $\max(\newmodel,\newmodel^*)$. We report mean values over three runs by varying the parameter $top\_p\in\{1.0,0.8,0.6\}$ and apply one tail Mann-Whitney U test for p-values. We observe a p-value of 0.05 while comparing the mean F1 score of \newmodel\ with \xinsta\ across languages and tasks.}
\label{tab:main}
\vspace{-3mm}
\end{table*}

\subsubsection{Training Dataset.}
A training dataset $D_t$ is constructed comprising pairs of sentences alongside their associated similarity scores. Specifically, $D_t$ consists of $(q_u^l,q_v^r \,|\, f(s)_{u,v})$, where $q_u^l$ denotes a low-resource query derived from $D_l$ with $u \in n$, $q_v^r$ indicates a query from a resource-rich dataset $D_r$ with $v \in m$, and $f(s)_{u,v}$ represents the similarity score between these queries within a low-resource monolingual embedding space.

Here, we present a systematic approach to construct $D_t$. Initially, we delineate ${D'}_r$, a transformed resource-rich dataset in a low-resource language using translation. Subsequently, we determine the temporal semantic alignment scores $f(s)$ among the queries in $D_l$ and ${D'}_r$. The utilization of all example pairs in the fine-tuning procedure incurs quadratic complexity in terms of $|D_t|$, rendering it resource-intensive. Drawing inspiration from \citet{rubin-etal-2022-learning}, this issue is addressed by selecting the top-$h$ analogous examples from ${D'}_r$ for each query in $D_l$. To counteract training data bias due to high similarity scores, we randomly select $w$ examples from the remaining dataset to capture the whole similarity distribution. Consequently, for every query $q_u^l \in D_l$, the resultant set $S_u$ comprises ($h+w$) sentence pairs, each accompanied by their temporal semantic similarity scores as postulated in Equation \ref{eq:pairset}.
\begin{equation}
\begin{split}
S_u  =  &\{(q_u^l,q_1^{r'}| f(s)_{u,1}), \dots, (q_u^l,q_h^{r'} | f(s)_{u,h}), \\ &(q_u^l,q_{h+1}^{r'} | f(s)_{u,h+1}), \dots, (q_u^l,q_{h+w}^{r'} | f(s)_{u,h+w}) \}
   \end{split}
   \label{eq:pairset}
\end{equation}
Likewise, paired sentences and their associated similarity scores are generated for all queries within $D_l$. In the concluding phase, the transformed, resource-rich dataset ${D'}_r$ is substituted back with the original dataset ${D}_r$. Specifically, the query $q_v^{r'}$, which represents a transformation of the resource-rich query $q_v^r$ into a low-resource language, is replaced with the original query $q_v^r$ within the paired sentences dataset. Multilingual Sentence-BERT \cite{reimers2019sentencebertsentenceembeddingsusing}, a pre-trained transformer model, is employed to derive the semantic alignment scores. 

\subsubsection{Fine-tuning the Retriever.}
CoSENT\footnote{https://kexue.fm/archives/8847} (Cosine Sentence) loss is employed for the fine-tuning of sentence pairs along with similarity scores as labels, utilizing multilingual Sentence-BERT as base retriever. CoSENT loss generates a more robust training signal for optimizing the cosine value than the traditional cosine similarity loss function. This loss function is shown in Equation \ref{eq:cosent}.
\begin{equation}
\begin{split}
\mathcal{L} = log \sum(1+&\exp(f(s)_{(q_a^l,q_b^r)}-f(s)_{(q_y^l,q_z^r)}+ \\ &\exp...)
   \end{split}
   \label{eq:cosent}
\end{equation}
In this context, where $(q_a^l,q_b^r)$ and $(q_y^l,q_z^r)$ represent instances from $D_t$ within a batch, under the condition that the anticipated similarity between $(a,b)$ exceeds that of $(y,z)$, the summation extends across all feasible input pairs within the batch satisfying this criterion. This approach amalgamates both cross-entropy and contrastive loss advantages. %This approach amalgamates the advantages of both cross-entropy and contrastive loss.

\section{Experimental Results And Analysis}
\label{section:results}
\subsection{Experimental Setup}

Primarily, we employ LLaMA3-8B \cite{llama3modelcard} for all experimental works. A three-shot ICL approach is used throughout the experimental setting, demonstrating superior outcomes compared to both one-shot and two-shot configurations. The value of $h$ and $w$ is set empirically at $30$ and $10$, respectively. To fine-tune the \newmodel\ retriever model, the `distiluse-base-multilingual-cased-v1' serves as the foundational model. This method is systematically applied to each low-resource language across temporal tasks -- L1, L2 and L3, to ensure optimum performance. Additionally, an integrated \newmodel\ retriever is fine-tuned across languages and temporal tasks. The Train and Dev datasets from \mtempreason\ are used to construct the parallel corpus to fine-tune the \newmodel\ retriever, with a separate held-out test set employed to benchmark all outcomes. We use word level F1 scores and exact match (EM) standards to quantify the LLM's responses. Please refer to the technical appendix for ablations on few-shots, parameters $h$ \& $w$, along with hyperparameters in detail.

\begin{table}[t]
\centering
\small
%\resizebox{.95\columnwidth}{!}{
\begin{tabular}{l|l|c|c|c}
\toprule
     & &\multicolumn{3}{c}{\textbf{F1.}} \\
    \cline{3-5}
    \textbf{Model} & \textbf{Method} & \textbf{L1} & \textbf{L2} & \textbf{L3}\\
    \hline
\multirow{2}*{{LLaMA3-8B}}& \xinsta$^\uparrow$ &46.62	&11.92	&17.74\\
& \newmodel & \textbf{57.15}&	\textbf{15.23}	&\textbf{19.87}\\
    \midrule
\multirow{2}*{{Mistral-v1}}& \xinsta$^\uparrow$ &38.26	&11.73&	\textbf{18.72}\\
& \newmodel & \textbf{46.45}&	\textbf{14.83}	&18.64\\
    \midrule
\multirow{2}*{{Vicuna-7b-v1.5}}& \xinsta$^\uparrow$ &27.93	&9.54	&12.68\\
& \newmodel & \textbf{36.67}	&\textbf{12.04}	&\textbf{12.89}\\
    \midrule
\multirow{2}*{{Bloomz-7b1}}& \xinsta$^\uparrow$ &29.45	&3.79	&13.49\\
& \newmodel &\textbf{40.74}	&\textbf{4.20}	&\textbf{13.57}\\
\midrule
&$\overline{\Delta}_{\newmodel-\uparrow}$ &\textcolor{blue}{9.68$\uparrow$}&\textcolor{blue}{2.33$\uparrow$}&\textcolor{blue}{$0.58\uparrow$}\\

\bottomrule
\end{tabular}
\caption{The performance of \newmodel\ across LLMs for temporal tasks using the French test set ($\Bar{\Delta}$: the mean improvement in F1 score across LLMs for a temporal task).} 
\label{tab:llm_gen}
\vspace{-5mm}
\end{table}

\subsection{\newmodel\ Advancements Over Precedence}
The comprehensive comparison of \newmodel\ with baselines highlights the effectiveness of incorporating cross-lingual time–sensitive semantic alignment compared to a conventional semantic aligner (\xinsta), as evidenced in Table \ref{tab:main} across a variety of low-resource languages and temporal tasks. The mean values of metrics were compared across three iterations by varying the model's parameter $top\_p$ (1.0, 0.8, 0.6). The parameter indicates the cumulative probability threshold for token selection. Notably, \newmodel\ achieves a mean increase of $5.32$, $1.62$, and $1.43$ points in the F1-score for French, German, and Romanian, respectively, with a p-value of $0.05$. Specifically, the most significant improvements in F1-score—$10.53$, $3.31$, and $2.13$ points for tasks L1, L2, and L3, respectively are observed in the French setting. A similar enhancement is discernible concerning the EM metric. Moreover, the overall analysis does not yield a definitive conclusion for an integrated \newmodel\ retriever over its counterpart except a notable transcend of $3.7$ points in the F1-score for task L1 within the German setting.

\subsection{Robustness Across \llms}

\begin{figure}[t]
\centering
\includegraphics[width=0.45\textwidth]{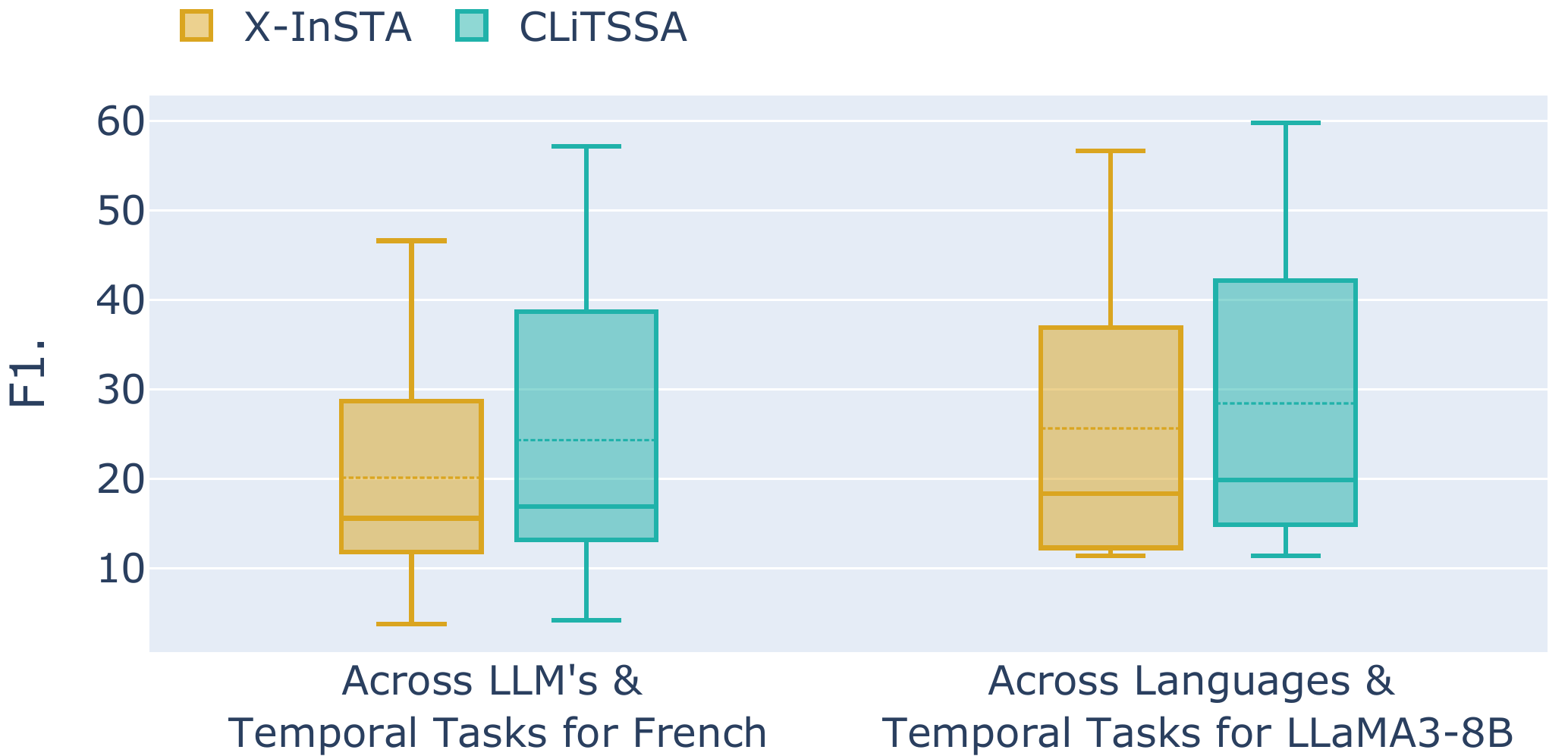}% Reduce the figure size so that it is slightly narrower than the column.
\caption{Comparison of F1 scores using box plot: a dual perspective on temporal tasks with language models and languages, pivoting on French and LLaMA3-8B, respectively.}
\label{fig:boxplot}
\vspace{-3mm}
\end{figure}

The assessment of \newmodel\ across a variety of distinct, contemporary LLMs namely, an English-dominant instruction-tuned model, Vicuna \cite{zheng2023judgingllmasajudgemtbenchchatbot}, a model fluent in French and German, Mistral \cite{jiang2023mistral7b}, and a cross-lingual specialized LLM, Bloomz \cite{muennighoff2023crosslingualgeneralizationmultitaskfinetuning}, demonstrates the robustness of the method. This evaluation highlights the versatility and effectiveness of \newmodel\ in engaging with and analyzing linguistic data across different languages and model architectures. The findings, as detailed in Table \ref{tab:llm_gen}, reveal that \newmodel\ surpasses the baseline by margins of $9.68$, $2.33$, and $0.58$ points in mean F1-score for L1, L2, and L3, respectively, across LLMs. 

To further substantiate the statistical significance of findings, a comparison is drawn through a box plot analysis of the F1-scores under two scenarios: firstly, by plotting F1-scores across LLMs and temporal tasks with a focus on the French language and secondly, through a box plot that contrasts F1-scores across various languages and temporal tasks centered around a specific LLM, namely LLaMA3-8B.  Figure \ref{fig:boxplot} shows that \newmodel\ notably extends the upper quartile by $10.01$ and $5.26$ points in terms of F1-score, with a mean increase of $4.20$ and $2.79$ points in the F1 scores, across these scenarios, respectively. Additionally, embedding space evolution under \newmodel, presented in technical appendix, further elucidates the noted enhancement.

\subsection{Cross-Task \newmodel\ Performance}
\begin{figure}[t]
\centering
\includegraphics[width=0.45\textwidth]{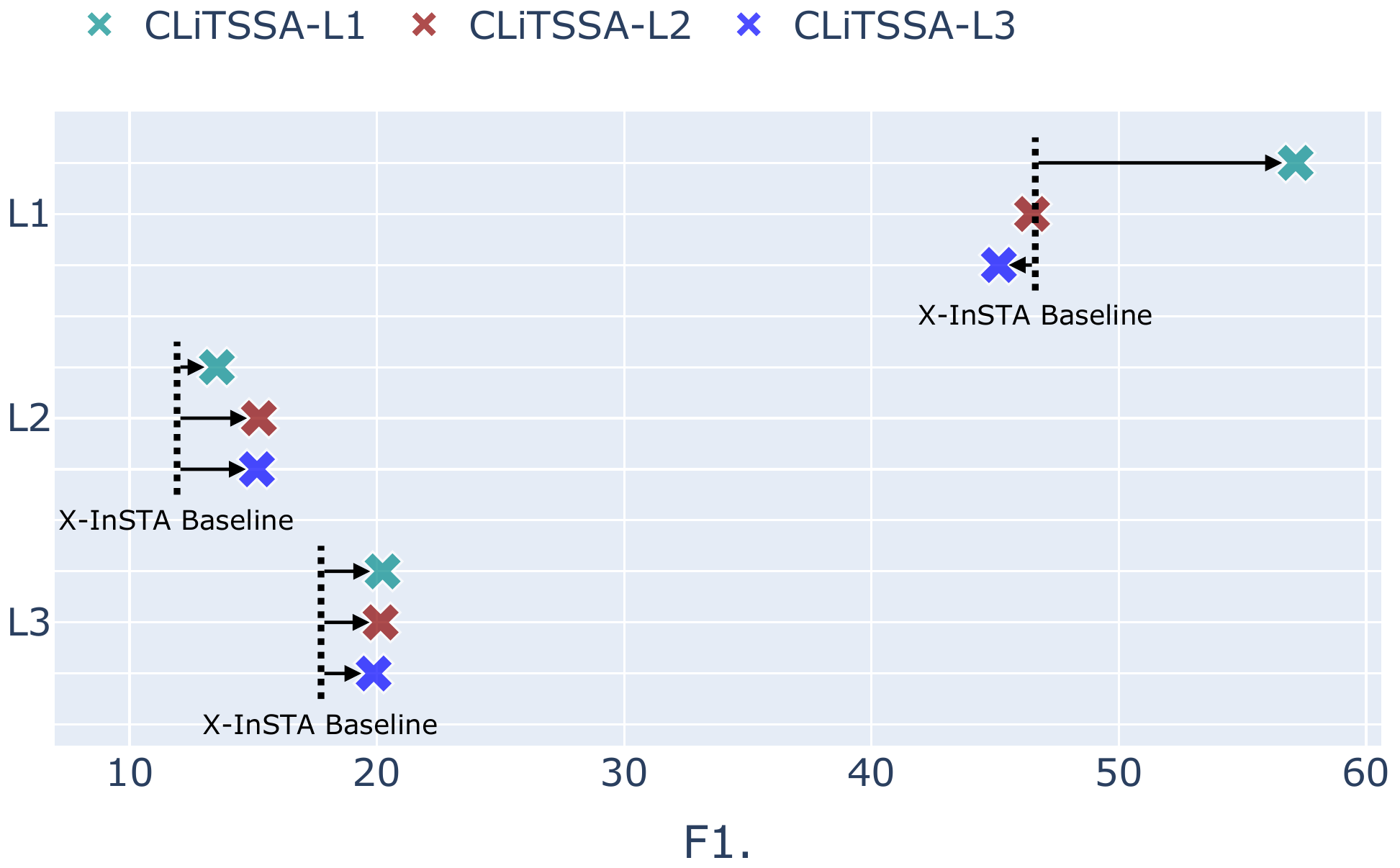} % Reduce the figure size so that it is slightly narrower than the column.
\caption{Cross–Task \newmodel\ performance across tasks with F1 scores on the French test set against the X-InSTA baseline. \newmodel-L$^*$ represents a retriever fine-tuned using L* training dataset where $* \in \{1,2,3\}$.}
\label{fig:crossperf}
\vspace{-3mm}
\end{figure}

Here, we evaluate the \newmodel\ retriever's generalization across temporal tasks to see if semantic alignment, sensitive to temporal variation achieved in one task, facilitates the resolution of another temporal task without necessitating a re-fine-tuning. The \newmodel\ model, once fine-tuned on a specific temporal task, is assessed on two other temporal tasks, i.e., the retriever optimized on the L1 task is employed to retrieve time–sensitive semantic examples for the L2 and L3 tasks. Figure \ref{fig:crossperf} presents the outcomes of this investigation. Note that tasks L1, L2, and L3 are sequentially arranged in order of temporal complexity, with L3 being the most intricate. The findings reveal that the temporal alignment acquired through the lower-level temporal task (L1) can significantly enhance the relative F1-score of the more complex tasks L2 and L3 by $13.5\%$ and $14.0\%$, respectively. However, the reverse scenario is inapplicable. Moreover, more complex tasks, L2 and L3, can exchange learning, thereby improving their F1-score relatively by $27.1\%$ and $13.6\%$, respectively. French is employed as the low-resource language in this study. The results corroborate that fine-tuning the \newmodel\ with a low-level temporal task (L1) could serve as a superior alternative to any semantic-based example retriever across temporal tasks.

\subsection{Cross-Linguality vs. Monolinguality}
\begin{figure}[t]
\centering
\includegraphics[width=0.45\textwidth]{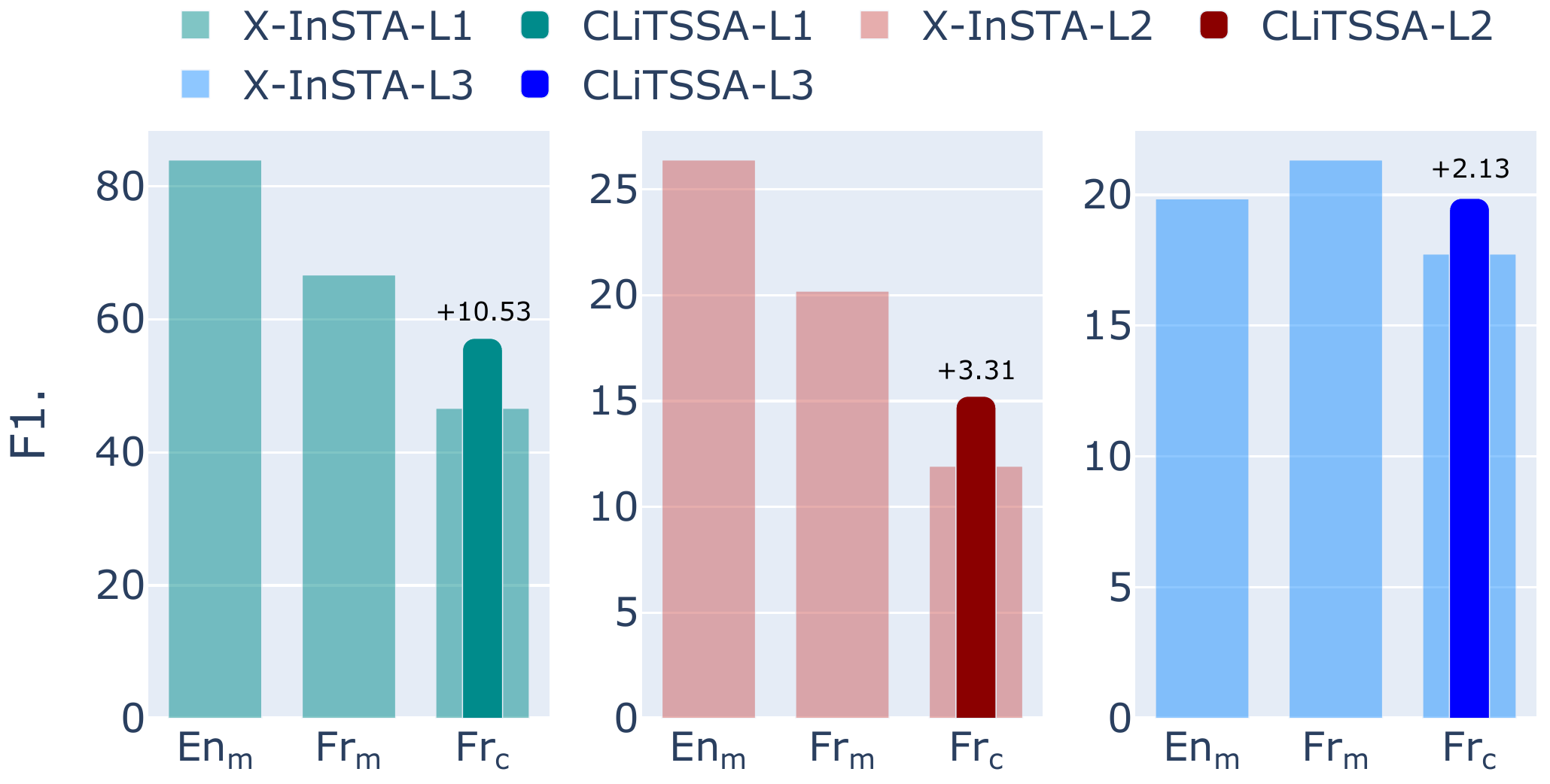} % Reduce the figure size so that it is slightly narrower than the column.
\caption{A comparative analysis of F1 scores across temporal tasks in monolingual and cross-lingual scenarios utilizing LLaMA3-8B, where \textit{En$_m$} and \textit{Fr$_m$} represent monolingual settings for English and French, respectively, while \textit{Fr$_c$} is French's cross-lingual setting.}
\label{fig:frmonocross}
\vspace{-4mm}
\end{figure}

The complexity of the prompt increases with the incorporation of multiple languages, which detrimentally impacts the performance of ICL in cross-lingual contexts when contrasted with monolingual scenarios. This experiment delineates the \newmodel's effectiveness in notably diminishing this discrepancy. As shown in Figure \ref{fig:frmonocross}, \newmodel\ ameliorates the performance differential between the French monolingual environment and cross-lingual context by $10.53$, $3.31$, and $2.13$ absolute points in terms of F1-score for the L1, L2, and L3 tasks, respectively, thereby comparing the results to those observed in a monolingual context. Contrastingly, the English monolingual scenario exhibits a significant divergence from its French counterpart for L1 and L2 tasks, underscoring the imperative for speedy enhancements to bolster performance in monolingual contexts for languages with limited resources.  

\begin{table*}[t]
\centering
\small
%\resizebox{.95\columnwidth}{!}{
\begin{tabular}{l|l|l|l}
\toprule
    \textbf{\#} & \textbf{Setup} & \textbf{Question/Answer} (\textbf{Q/A})&\textbf{Predicted Answer} \\
    \hline
\multirow{2}*{{1}}& X, En$_m$ & \textbf{Q:} Which team did Glynn Snodin play for in Oct, 1986?, \textbf{A:} Sheffield Wednesday F.C. &\cellcolor{LightRed} {Leeds United F.C.}\\
& C, Fr$_c$ &\textbf{Q:} Pour quelle équipe a joué Glynn Snodin en octobre 1986?, \textbf{A:} Sheffield Wednesday F.C. &\cellcolor{LightRed} {Leeds United}\\
%	&&&\\
\midrule
\multirow{2}*{{2}}& X, Fr$_m$ &\textbf{Q:} Qui était l'entraîneur principal de l'équipe HNK Rijeka après Simon Roman?, \textbf{A:} Goran Tomi &\cellcolor{LightRed} {Ivan Pudar}\\
& C, Fr$_c$ &\textbf{Q:} Qui était l'entraîneur principal de l'équipe HNK Rijeka après Simon Roman?, \textbf{A:} Goran Tomi &\cellcolor{LightRed} {Ivan Pudar}\\
%&  & & \textcolor{gray}{(Translation: Who was the head coach of HNK Rijeka team after Simon Roman?)} &	 &\\

\bottomrule
\end{tabular}
\caption{Failure cases with \newmodel\ and their corresponding responses from \xinsta\ in the monolingual scenario. \textit{En$_m$}: English monolingual, \textit{Fr$_m$}: French monolingual, and \textit{Fr$_c$}: French cross-lingual. X: \xinsta, and C: \newmodel}
\label{tab:errorana}
\vspace{-2mm}
\end{table*}

\section{Error Analysis}
\label{section:error}
Table \ref{tab:errorana} shows a couple of instances underscoring the challenges of semantic alignment. Initially, the heightened time–sensitive alignment offered by our model does not rectify the \textbf{inaccuracies in the foundational knowledge} of the underlying LLM. The first example elucidates that the factual inaccuracies inherent in the LLaMA3-8B model within a resource-rich linguistic context (i.e., English monolingual, $En_m$) persist despite the application of \newmodel\ in a French cross-lingual setting ($Fr_c$). Additionally, our proposed methodology is contingent on the semantic context within the monolingual embedding space, aligning the cross-lingual space accordingly. Consequently, \textbf{inaccuracies in expected responses may propagate} from the monolingual to the cross-lingual space notwithstanding the enhanced query alignment. A subsequent example illustrates this phenomenon in the contexts of French monolingual ($Fr_m$) and French cross-lingual scenarios. Furthermore, notwithstanding the semantic alignment ingrained in cross-lingual queries, the \textbf{implicit aspect of temporality} persistently presents a challenge, as observed for the L3 task.
\section{Related Works}
\label{section:literature}
In NLP, significant foundational efforts in temporal reasoning encompass the creation of TimeBank \cite{article}, TempEval \cite{verhagen-etal-2010-semeval}, and Time–stamped Language Models \cite{rajaby-faghihi-kordjamshidi-2021-time}, each contributing significantly to the understanding and processing temporal data. Concurrently, the evolution of knowledge graphs (KGs) has accentuated the importance of temporal relations therein, catalyzing the emergence of Temporal Knowledge Graph Completion (TKGC) as a distinct area of study. This progression has given rise to noteworthy question-answering datasets predicated on TKG, including TEQUILA \cite{Jia_2018}, TimeQuestions \cite{Jia_2021}, and CronQuestions \cite{saxena-etal-2021-question}. The widespread use of language models in the public sphere further underscores the necessity for both temporal accuracy and consistency within generated responses. In response to this demand, several time–sensitive QA datasets, such as TEMPLAMA \cite{dhingra-etal-2022-time} and TEMPREASON \cite{tan2023benchmarkingimprovingtemporalreasoning}, have been introduced to assess and benchmark the temporal reasoning capabilities inherent in \llms. Among these, TEMPREASON stands out as a comprehensive benchmark for temporal reasoning, spanning a broad spectrum of temporal periods and incorporating three levels of temporal relations. Further, the TEMP-COFAC \cite{bajpai-etal-2024-temporally} has been introduced to assess the temporally consistent factuality.

Furthermore, most \llms\ are trained on multilingual datasets \cite{wenzek-etal-2020-ccnet}, a practice that was once a rarity given the dominance of extensive English corpora \cite{Radford2019LanguageMA}. Despite this, \llms\ have proven their mettle in considerable languages. While there have been significant advancements in the multilingual capabilities of \llms\ \cite{lin-etal-2022-shot,qin2024multilinguallargelanguagemodel}, they still face significant challenges when dealing with low-resource languages \cite{cahyawijaya2024llmsfewshotincontextlowresource}, especially in task-specific contexts \cite{enis2024llmnmtadvancinglowresource}. To address this, innovative approaches such as prompting for generating intermediate English contexts \cite{huang2023languagescreatedequalllms}, cross-lingual prompting \cite{winata-etal-2021-language}, and Linguistically Diverse Prompting (LDP) \cite{nguyen2024democratizingllmslowresourcelanguages} have been introduced. In the cross-lingual prompting domain, specific developments like semantic label-based alignment \cite{tanwar2023multilingualllmsbettercrosslingual}, query-based alignment via translation semantic similarity \cite{cahyawijaya2024llmsfewshotincontextlowresource}, and a model-specific fine-tuned retriever \cite{lin2024xamplerlearningretrievecrosslingual} have further enhanced LLMs' capabilities. Yet, the exploration of temporal reasoning within low-resource languages remains scant, presenting a compelling area for further research. This study proposes to pioneer advancements in this under-explored domain.
\section{Conclusion}
\label{section:conclusion}
In this paper, we introduced a novel dataset, \mtempreason, aimed at improving temporal reasoning assessment in low-resource languages using LLMs. Our analyses identified that multilingual LLMs inherently reward in-language time–sensitive semantic alignment over the cross-lingual similarity context in the X-ICL method. To overcome this, we proposed \newmodel, a novel method that enhances the retrieval of time–sensitive contextually relevant examples across low-resource languages. Our results demonstrated that this approach effectively improves \llms' temporal reasoning in low-resource languages, which we believe will aid in promoting linguistic diversity and the development of more inclusive \llms. Future endeavors may benefit from examining the alignment between an implicit temporal query's semantics and its implied semantic space to enhance intricate L3 task performance.

\section{Acknowledgments}
T. Chakraborty  acknowledges the support of
the IBM-IITD AI Horizons network and Rajiv
Khemani Young Faculty Chair Professorship in Artificial Intelligence.

\bibliography{paper}

\begin{thebibliography}{37}
\providecommand{\natexlab}[1]{#1}

\bibitem[{Adilazuarda et~al.(2024)Adilazuarda, Cahyawijaya, Aji, Winata, and Purwarianti}]{adilazuarda2024lingualchemyfusingtypologicalgeographical}
Adilazuarda, M.~F.; Cahyawijaya, S.; Aji, A.~F.; Winata, G.~I.; and Purwarianti, A. 2024.
\newblock LinguAlchemy: Fusing Typological and Geographical Elements for Unseen Language Generalization.
\newblock arXiv:2401.06034.

\bibitem[{AI@Meta(2024)}]{llama3modelcard}
AI@Meta. 2024.
\newblock Llama 3 Model Card.

\bibitem[{Asai et~al.(2023)Asai, Kudugunta, Yu, Blevins, Gonen, Reid, Tsvetkov, Ruder, and Hajishirzi}]{asai2023buffetbenchmarkinglargelanguage}
Asai, A.; Kudugunta, S.; Yu, X.~V.; Blevins, T.; Gonen, H.; Reid, M.; Tsvetkov, Y.; Ruder, S.; and Hajishirzi, H. 2023.
\newblock BUFFET: Benchmarking Large Language Models for Few-shot Cross-lingual Transfer.
\newblock arXiv:2305.14857.

\bibitem[{Bajpai et~al.(2024)Bajpai, Goyal, Anwer, and Chakraborty}]{bajpai-etal-2024-temporally}
Bajpai, A.; Goyal, A.; Anwer, A.; and Chakraborty, T. 2024.
\newblock Temporally Consistent Factuality Probing for Large Language Models.
\newblock In Al-Onaizan, Y.; Bansal, M.; and Chen, Y.-N., eds., \emph{Proceedings of the 2024 Conference on Empirical Methods in Natural Language Processing}, 15864--15881. Miami, Florida, USA: Association for Computational Linguistics.

\bibitem[{Cahyawijaya et~al.(2023)Cahyawijaya, Lovenia, Aji, Winata, Wilie, Mahendra, Wibisono, Romadhony, Vincentio, Koto, Santoso, Moeljadi, Wirawan, Hudi, Parmonangan, Alfina, Wicaksono, Putra, Rahmadani, Oenang, Septiandri, Jaya, Dhole, Suryani, Putri, Su, Stevens, Nityasya, Adilazuarda, Ignatius, Diandaru, Yu, Ghifari, Dai, Xu, Damapuspita, Tho, Karo, Fatyanosa, Ji, Fung, Neubig, Baldwin, Ruder, Sujaini, Sakti, and Purwarianti}]{cahyawijaya2023nusacrowdopensourceinitiative}
Cahyawijaya, S.; Lovenia, H.; Aji, A.~F.; Winata, G.~I.; Wilie, B.; Mahendra, R.; Wibisono, C.; Romadhony, A.; Vincentio, K.; Koto, F.; Santoso, J.; Moeljadi, D.; Wirawan, C.; Hudi, F.; Parmonangan, I.~H.; Alfina, I.; Wicaksono, M.~S.; Putra, I.~F.; Rahmadani, S.; Oenang, Y.; Septiandri, A.~A.; Jaya, J.; Dhole, K.~D.; Suryani, A.~A.; Putri, R.~A.; Su, D.; Stevens, K.; Nityasya, M.~N.; Adilazuarda, M.~F.; Ignatius, R.; Diandaru, R.; Yu, T.; Ghifari, V.; Dai, W.; Xu, Y.; Damapuspita, D.; Tho, C.; Karo, I. M.~K.; Fatyanosa, T.~N.; Ji, Z.; Fung, P.; Neubig, G.; Baldwin, T.; Ruder, S.; Sujaini, H.; Sakti, S.; and Purwarianti, A. 2023.
\newblock NusaCrowd: Open Source Initiative for Indonesian NLP Resources.
\newblock arXiv:2212.09648.

\bibitem[{Cahyawijaya, Lovenia, and Fung(2024)}]{cahyawijaya2024llmsfewshotincontextlowresource}
Cahyawijaya, S.; Lovenia, H.; and Fung, P. 2024.
\newblock LLMs Are Few-Shot In-Context Low-Resource Language Learners.
\newblock arXiv:2403.16512.

\bibitem[{Chen, Wang, and Wang(2021)}]{chen2021datasetansweringtimesensitivequestions}
Chen, W.; Wang, X.; and Wang, W.~Y. 2021.
\newblock A Dataset for Answering Time-Sensitive Questions.
\newblock arXiv:2108.06314.

\bibitem[{Dhingra et~al.(2022)Dhingra, Cole, Eisenschlos, Gillick, Eisenstein, and Cohen}]{dhingra-etal-2022-time}
Dhingra, B.; Cole, J.~R.; Eisenschlos, J.~M.; Gillick, D.; Eisenstein, J.; and Cohen, W.~W. 2022.
\newblock Time-Aware Language Models as Temporal Knowledge Bases.
\newblock \emph{Transactions of the Association for Computational Linguistics}, 10: 257--273.

\bibitem[{Enis and Hopkins(2024)}]{enis2024llmnmtadvancinglowresource}
Enis, M.; and Hopkins, M. 2024.
\newblock From LLM to NMT: Advancing Low-Resource Machine Translation with Claude.
\newblock arXiv:2404.13813.

\bibitem[{Huang et~al.(2023)Huang, Tang, Zhang, Zhao, Song, Xia, and Wei}]{huang2023languagescreatedequalllms}
Huang, H.; Tang, T.; Zhang, D.; Zhao, W.~X.; Song, T.; Xia, Y.; and Wei, F. 2023.
\newblock Not All Languages Are Created Equal in LLMs: Improving Multilingual Capability by Cross-Lingual-Thought Prompting.
\newblock arXiv:2305.07004.

\bibitem[{Jia et~al.(2018)Jia, Abujabal, Saha~Roy, Strötgen, and Weikum}]{Jia_2018}
Jia, Z.; Abujabal, A.; Saha~Roy, R.; Strötgen, J.; and Weikum, G. 2018.
\newblock TEQUILA: Temporal Question Answering over Knowledge Bases.
\newblock In \emph{Proceedings of the 27th ACM International Conference on Information and Knowledge Management}, CIKM ’18. ACM.

\bibitem[{Jia et~al.(2021)Jia, Pramanik, Saha~Roy, and Weikum}]{Jia_2021}
Jia, Z.; Pramanik, S.; Saha~Roy, R.; and Weikum, G. 2021.
\newblock Complex Temporal Question Answering on Knowledge Graphs.
\newblock In \emph{Proceedings of the 30th ACM International Conference on Information and Knowledge Management}, CIKM ’21. ACM.

\bibitem[{Jiang et~al.(2023)Jiang, Sablayrolles, Mensch, Bamford, Chaplot, de~las Casas, Bressand, Lengyel, Lample, Saulnier, Lavaud, Lachaux, Stock, Scao, Lavril, Wang, Lacroix, and Sayed}]{jiang2023mistral7b}
Jiang, A.~Q.; Sablayrolles, A.; Mensch, A.; Bamford, C.; Chaplot, D.~S.; de~las Casas, D.; Bressand, F.; Lengyel, G.; Lample, G.; Saulnier, L.; Lavaud, L.~R.; Lachaux, M.-A.; Stock, P.; Scao, T.~L.; Lavril, T.; Wang, T.; Lacroix, T.; and Sayed, W.~E. 2023.
\newblock Mistral 7B.
\newblock arXiv:2310.06825.

\bibitem[{Lin, Martins, and Schütze(2024)}]{lin2024xamplerlearningretrievecrosslingual}
Lin, P.; Martins, A. F.~T.; and Schütze, H. 2024.
\newblock XAMPLER: Learning to Retrieve Cross-Lingual In-Context Examples.
\newblock arXiv:2405.05116.

\bibitem[{Lin et~al.(2022{\natexlab{a}})Lin, Mihaylov, Artetxe, Wang, Chen, Simig, Ott, Goyal, Bhosale, Du, Pasunuru, Shleifer, Koura, Chaudhary, O{'}Horo, Wang, Zettlemoyer, Kozareva, Diab, Stoyanov, and Li}]{lin-etal-2022-shot}
Lin, X.~V.; Mihaylov, T.; Artetxe, M.; Wang, T.; Chen, S.; Simig, D.; Ott, M.; Goyal, N.; Bhosale, S.; Du, J.; Pasunuru, R.; Shleifer, S.; Koura, P.~S.; Chaudhary, V.; O{'}Horo, B.; Wang, J.; Zettlemoyer, L.; Kozareva, Z.; Diab, M.; Stoyanov, V.; and Li, X. 2022{\natexlab{a}}.
\newblock Few-shot Learning with Multilingual Generative Language Models.
\newblock In Goldberg, Y.; Kozareva, Z.; and Zhang, Y., eds., \emph{Proceedings of the 2022 Conference on Empirical Methods in Natural Language Processing}, 9019--9052. Abu Dhabi, United Arab Emirates: Association for Computational Linguistics.

\bibitem[{Lin et~al.(2022{\natexlab{b}})Lin, Mihaylov, Artetxe, Wang, Chen, Simig, Ott, Goyal, Bhosale, Du, Pasunuru, Shleifer, Koura, Chaudhary, O'Horo, Wang, Zettlemoyer, Kozareva, Diab, Stoyanov, and Li}]{lin2022fewshotlearningmultilinguallanguage}
Lin, X.~V.; Mihaylov, T.; Artetxe, M.; Wang, T.; Chen, S.; Simig, D.; Ott, M.; Goyal, N.; Bhosale, S.; Du, J.; Pasunuru, R.; Shleifer, S.; Koura, P.~S.; Chaudhary, V.; O'Horo, B.; Wang, J.; Zettlemoyer, L.; Kozareva, Z.; Diab, M.; Stoyanov, V.; and Li, X. 2022{\natexlab{b}}.
\newblock Few-shot Learning with Multilingual Language Models.
\newblock arXiv:2112.10668.

\bibitem[{Liu et~al.(2022)Liu, Shen, Zhang, Dolan, Carin, and Chen}]{liu-etal-2022-makes}
Liu, J.; Shen, D.; Zhang, Y.; Dolan, B.; Carin, L.; and Chen, W. 2022.
\newblock What Makes Good In-Context Examples for {GPT}-3?
\newblock In Agirre, E.; Apidianaki, M.; and Vuli{\'c}, I., eds., \emph{Proceedings of Deep Learning Inside Out (DeeLIO 2022): The 3rd Workshop on Knowledge Extraction and Integration for Deep Learning Architectures}, 100--114. Dublin, Ireland and Online: Association for Computational Linguistics.

\bibitem[{Miyabe and Yoshino(2015)}]{Miyabe2015EvaluationOT}
Miyabe, M.; and Yoshino, T. 2015.
\newblock Evaluation of the Validity of Back-Translation as a Method of Assessing the Accuracy of Machine Translation.
\newblock \emph{2015 International Conference on Culture and Computing (Culture Computing)}, 145--150.

\bibitem[{Muennighoff et~al.(2023)Muennighoff, Wang, Sutawika, Roberts, Biderman, Scao, Bari, Shen, Yong, Schoelkopf, Tang, Radev, Aji, Almubarak, Albanie, Alyafeai, Webson, Raff, and Raffel}]{muennighoff2023crosslingualgeneralizationmultitaskfinetuning}
Muennighoff, N.; Wang, T.; Sutawika, L.; Roberts, A.; Biderman, S.; Scao, T.~L.; Bari, M.~S.; Shen, S.; Yong, Z.-X.; Schoelkopf, H.; Tang, X.; Radev, D.; Aji, A.~F.; Almubarak, K.; Albanie, S.; Alyafeai, Z.; Webson, A.; Raff, E.; and Raffel, C. 2023.
\newblock Crosslingual Generalization through Multitask Finetuning.
\newblock arXiv:2211.01786.

\bibitem[{Nguyen et~al.(2024)Nguyen, Aljunied, Joty, and Bing}]{nguyen2024democratizingllmslowresourcelanguages}
Nguyen, X.-P.; Aljunied, S.~M.; Joty, S.; and Bing, L. 2024.
\newblock Democratizing LLMs for Low-Resource Languages by Leveraging their English Dominant Abilities with Linguistically-Diverse Prompts.
\newblock arXiv:2306.11372.

\bibitem[{Pustejovsky et~al.(2003)Pustejovsky, Hanks, Saurí, See, Gaizauskas, Setzer, Radev, Sundheim, Day, Ferro, and Lazo}]{article}
Pustejovsky, J.; Hanks, P.; Saurí, R.; See, A.; Gaizauskas, R.; Setzer, A.; Radev, D.; Sundheim, B.; Day, D.; Ferro, L.; and Lazo, M. 2003.
\newblock The TimeBank corpus.
\newblock \emph{Proceedings of Corpus Linguistics}.

\bibitem[{Qin et~al.(2024)Qin, Chen, Zhou, Chen, Li, Liao, Li, Che, and Yu}]{qin2024multilinguallargelanguagemodel}
Qin, L.; Chen, Q.; Zhou, Y.; Chen, Z.; Li, Y.; Liao, L.; Li, M.; Che, W.; and Yu, P.~S. 2024.
\newblock Multilingual Large Language Model: A Survey of Resources, Taxonomy and Frontiers.
\newblock arXiv:2404.04925.

\bibitem[{Radford et~al.(2019)Radford, Wu, Child, Luan, Amodei, and Sutskever}]{Radford2019LanguageMA}
Radford, A.; Wu, J.; Child, R.; Luan, D.; Amodei, D.; and Sutskever, I. 2019.
\newblock Language Models are Unsupervised Multitask Learners.

\bibitem[{Raffel et~al.(2023)Raffel, Shazeer, Roberts, Lee, Narang, Matena, Zhou, Li, and Liu}]{raffel2023exploringlimitstransferlearning}
Raffel, C.; Shazeer, N.; Roberts, A.; Lee, K.; Narang, S.; Matena, M.; Zhou, Y.; Li, W.; and Liu, P.~J. 2023.
\newblock Exploring the Limits of Transfer Learning with a Unified Text-to-Text Transformer.
\newblock arXiv:1910.10683.

\bibitem[{Rajaby~Faghihi and Kordjamshidi(2021)}]{rajaby-faghihi-kordjamshidi-2021-time}
Rajaby~Faghihi, H.; and Kordjamshidi, P. 2021.
\newblock Time-Stamped Language Model: Teaching Language Models to Understand The Flow of Events.
\newblock In Toutanova, K.; Rumshisky, A.; Zettlemoyer, L.; Hakkani-Tur, D.; Beltagy, I.; Bethard, S.; Cotterell, R.; Chakraborty, T.; and Zhou, Y., eds., \emph{Proceedings of the 2021 Conference of the North American Chapter of the Association for Computational Linguistics: Human Language Technologies}, 4560--4570. Online: Association for Computational Linguistics.

\bibitem[{Reimers and Gurevych(2019)}]{reimers2019sentencebertsentenceembeddingsusing}
Reimers, N.; and Gurevych, I. 2019.
\newblock Sentence-BERT: Sentence Embeddings using Siamese BERT-Networks.
\newblock arXiv:1908.10084.

\bibitem[{Rubin, Herzig, and Berant(2022)}]{rubin-etal-2022-learning}
Rubin, O.; Herzig, J.; and Berant, J. 2022.
\newblock Learning To Retrieve Prompts for In-Context Learning.
\newblock In Carpuat, M.; de~Marneffe, M.-C.; and Meza~Ruiz, I.~V., eds., \emph{Proceedings of the 2022 Conference of the North American Chapter of the Association for Computational Linguistics: Human Language Technologies}, 2655--2671. Seattle, United States: Association for Computational Linguistics.

\bibitem[{Saxena, Chakrabarti, and Talukdar(2021)}]{saxena-etal-2021-question}
Saxena, A.; Chakrabarti, S.; and Talukdar, P. 2021.
\newblock Question Answering Over Temporal Knowledge Graphs.
\newblock In Zong, C.; Xia, F.; Li, W.; and Navigli, R., eds., \emph{Proceedings of the 59th Annual Meeting of the Association for Computational Linguistics and the 11th International Joint Conference on Natural Language Processing (Volume 1: Long Papers)}, 6663--6676. Online: Association for Computational Linguistics.

\bibitem[{Tan, Ng, and Bing(2023)}]{tan2023benchmarkingimprovingtemporalreasoning}
Tan, Q.; Ng, H.~T.; and Bing, L. 2023.
\newblock Towards Benchmarking and Improving the Temporal Reasoning Capability of Large Language Models.
\newblock arXiv:2306.08952.

\bibitem[{Tanwar et~al.(2023)Tanwar, Dutta, Borthakur, and Chakraborty}]{tanwar2023multilingualllmsbettercrosslingual}
Tanwar, E.; Dutta, S.; Borthakur, M.; and Chakraborty, T. 2023.
\newblock Multilingual LLMs are Better Cross-lingual In-context Learners with Alignment.
\newblock arXiv:2305.05940.

\bibitem[{Verhagen et~al.(2010)Verhagen, Saur{\'\i}, Caselli, and Pustejovsky}]{verhagen-etal-2010-semeval}
Verhagen, M.; Saur{\'\i}, R.; Caselli, T.; and Pustejovsky, J. 2010.
\newblock {S}em{E}val-2010 Task 13: {T}emp{E}val-2.
\newblock In Erk, K.; and Strapparava, C., eds., \emph{Proceedings of the 5th International Workshop on Semantic Evaluation}, 57--62. Uppsala, Sweden: Association for Computational Linguistics.

\bibitem[{Wenzek et~al.(2020)Wenzek, Lachaux, Conneau, Chaudhary, Guzm{\'a}n, Joulin, and Grave}]{wenzek-etal-2020-ccnet}
Wenzek, G.; Lachaux, M.-A.; Conneau, A.; Chaudhary, V.; Guzm{\'a}n, F.; Joulin, A.; and Grave, E. 2020.
\newblock {CCN}et: Extracting High Quality Monolingual Datasets from Web Crawl Data.
\newblock In Calzolari, N.; B{\'e}chet, F.; Blache, P.; Choukri, K.; Cieri, C.; Declerck, T.; Goggi, S.; Isahara, H.; Maegaard, B.; Mariani, J.; Mazo, H.; Moreno, A.; Odijk, J.; and Piperidis, S., eds., \emph{Proceedings of the Twelfth Language Resources and Evaluation Conference}, 4003--4012. Marseille, France: European Language Resources Association.
\newblock ISBN 979-10-95546-34-4.

\bibitem[{Winata et~al.(2021)Winata, Madotto, Lin, Liu, Yosinski, and Fung}]{winata-etal-2021-language}
Winata, G.~I.; Madotto, A.; Lin, Z.; Liu, R.; Yosinski, J.; and Fung, P. 2021.
\newblock Language Models are Few-shot Multilingual Learners.
\newblock In Ataman, D.; Birch, A.; Conneau, A.; Firat, O.; Ruder, S.; and Sahin, G.~G., eds., \emph{Proceedings of the 1st Workshop on Multilingual Representation Learning}, 1--15. Punta Cana, Dominican Republic: Association for Computational Linguistics.

\bibitem[{Yamada and Ri(2024)}]{yamada2024leiafacilitatingcrosslingualknowledge}
Yamada, I.; and Ri, R. 2024.
\newblock LEIA: Facilitating Cross-lingual Knowledge Transfer in Language Models with Entity-based Data Augmentation.
\newblock arXiv:2402.11485.

\bibitem[{Zhang et~al.(2022)Zhang, Li, Chen, Deng, Bi, Tan, Huang, and Chen}]{zhang2022differentiablepromptmakespretrained}
Zhang, N.; Li, L.; Chen, X.; Deng, S.; Bi, Z.; Tan, C.; Huang, F.; and Chen, H. 2022.
\newblock Differentiable Prompt Makes Pre-trained Language Models Better Few-shot Learners.
\newblock arXiv:2108.13161.

\bibitem[{Zhao et~al.(2021)Zhao, Wallace, Feng, Klein, and Singh}]{zhao2021calibrateuseimprovingfewshot}
Zhao, T.~Z.; Wallace, E.; Feng, S.; Klein, D.; and Singh, S. 2021.
\newblock Calibrate Before Use: Improving Few-Shot Performance of Language Models.
\newblock arXiv:2102.09690.

\bibitem[{Zheng et~al.(2023)Zheng, Chiang, Sheng, Zhuang, Wu, Zhuang, Lin, Li, Li, Xing, Zhang, Gonzalez, and Stoica}]{zheng2023judgingllmasajudgemtbenchchatbot}
Zheng, L.; Chiang, W.-L.; Sheng, Y.; Zhuang, S.; Wu, Z.; Zhuang, Y.; Lin, Z.; Li, Z.; Li, D.; Xing, E.~P.; Zhang, H.; Gonzalez, J.~E.; and Stoica, I. 2023.
\newblock Judging LLM-as-a-Judge with MT-Bench and Chatbot Arena.
\newblock arXiv:2306.05685.

\end{thebibliography}

\appendix

\section{Technical Appendix}

In this section, we present a few more supplementary evidence in support of \newmodel. Specifically, the evolution of embedding space under \newmodel, extended experimental results, and outlining hyperparameters. 

\subsection{Cross-Lingual In-Context Few-Shot Performance}

In this experiment, we introduce an ablation concerning the k-shot cross-lingual in-context \xinsta\ performance spanning temporal tasks, specifically focusing on French as a low-resource language on LLaMA-3 [8B], where $k \in \{1,2,3\}$. The outcomes delineated in Table \ref{tab:icl} demonstrate superior performance in a three-shot scenario with reaching at a saturation level. This finding led to the selection of this configuration across all experiments in this work.

\begin{table}[h]
\centering
%\resizebox{.95\columnwidth}{!}{
\begin{tabular}{l|l|c|c|c}
\toprule
    \textbf{Metric} & \textbf{Task} & \textbf{1-shot} & \textbf{2-shot}& \textbf{3-shot}  \\
    \midrule

\multirow{3}*{{F1.}} & L1 & 21.46	&44.77	&\textbf{46.62} \\
& L2 & 8.04	&11.56&	\textbf{11.92} \\
& L3 & 17.03&	16.63&	\textbf{17.74}\\
\midrule
\multirow{3}*{{EM}} & L1 & 9.2&20.77	&\textbf{22.05}\\
& L2 &0.44	&4.28	&\textbf{4.55} \\
& L3 & 9.19	&9.96	&\textbf{10.55}\\

\bottomrule
\end{tabular}
\caption{A few-shot comparison of F1 scores and EM metrics for cross-lingual semantically aligned in-context learning in French settings across temporal tasks.}
\label{tab:icl}
\end{table}

\subsection{Ablation For Parameters $h$ and $w$}

In this section, the emphasis lies on identifying the optimal values for parameters $h$ and $w$, which are pivotal in constructing the training data for \newmodel\ fine-tuning. This exploration is performed using the French dataset for experimental ablation. The determination of the optimal values for  $h$ and $w$ is guided by three primary objectives: Firstly, to minimize the divergence in similarity contexts between the subsample space and the entire sample space; Secondly, to emphasize the similarity contexts for positive pairs more than for other pairs, given that the goal is to retrieve the fine-grained top-k semantically similar examples—this emphasis assists in learning the nuanced differences among positive semantic contexts. Lastly, to adhere to a limitation on the overall size of the training sample. 

\begin{table}[h]
\centering
%\resizebox{.95\columnwidth}{!}{
\begin{tabular}{l|c|c|c}
\toprule
 \textbf{h / w} & \textbf{5} & \textbf{10}& \textbf{15}  \\
    \midrule
\multicolumn{3}{l}{{KL Divergence}}\\

\textbf{20} &0.12&	\cellcolor{LighterGreen}{0.08}&	\cellcolor{LighterGreen}{0.06}\\
\textbf{30} &0.14&	\cellcolor{LighterGreen}{0.10}&	\cellcolor{LighterGreen}{0.07}\\
\textbf{40} & 0.16&	0.11&	\cellcolor{LighterGreen}{0.09}\\
\midrule
\multicolumn{3}{l}{{Prioritization Factor}}\\
\textbf{20}&0.8	&\cellcolor{LighterGreen}{0.66}&	\cellcolor{LighterGreen}{0.57}\\
\textbf{30} &0.85&	\cellcolor{LightGreen}{0.75}&	\cellcolor{LighterGreen}{0.66}\\
\textbf{40} &0.88&	0.8	&\cellcolor{LightGreen}{0.72}\\
\midrule
\multicolumn{3}{l}{{Training Sample Size}}\\
\textbf{20}&25000&	\cellcolor{LighterGreen}{30000}&	\cellcolor{LighterGreen}{40000}\\
\textbf{30} &35000&	\cellcolor{Green}{40000}&	\cellcolor{LighterGreen}{45000}\\
\textbf{40} &45000&	50000&	\cellcolor{LightGreen}{55000}\\

\bottomrule
\end{tabular}
\caption{Ablation for parameters h and w. Iterative filtering in three stages to yield optimal values.}
\label{tab:hnw}
\end{table}

To measure the divergence between the semantic similarity distributions of subsample spaces—created by altering the value of $h \in \{20,30,40\}$ and $w \in \{5,10,15\}$—and the distribution of the entire sample space, the KL-Divergence metric is utilized. The prioritization factor is denoted as $h/(h+w)$. Table \ref{tab:hnw} delineates the empirical iterative process employed for identifying the optimal values of  $h$ and $w$. Initially, the top-5 combinations characterized by minimal divergence are selected; from these, the top-2 combinations where the prioritization factor is maximized are chosen. The final selection is the combination that results in the lowest training sample size. This methodological approach leads to the determination of $h=30$ and $w=10$ as the optimal parameters, which were then adopted for the main experiments.

\subsection{Evolution Of Embedding Space Under \newmodel}

\begin{figure}[h!]
\centering
\includegraphics[width=0.4\textwidth]{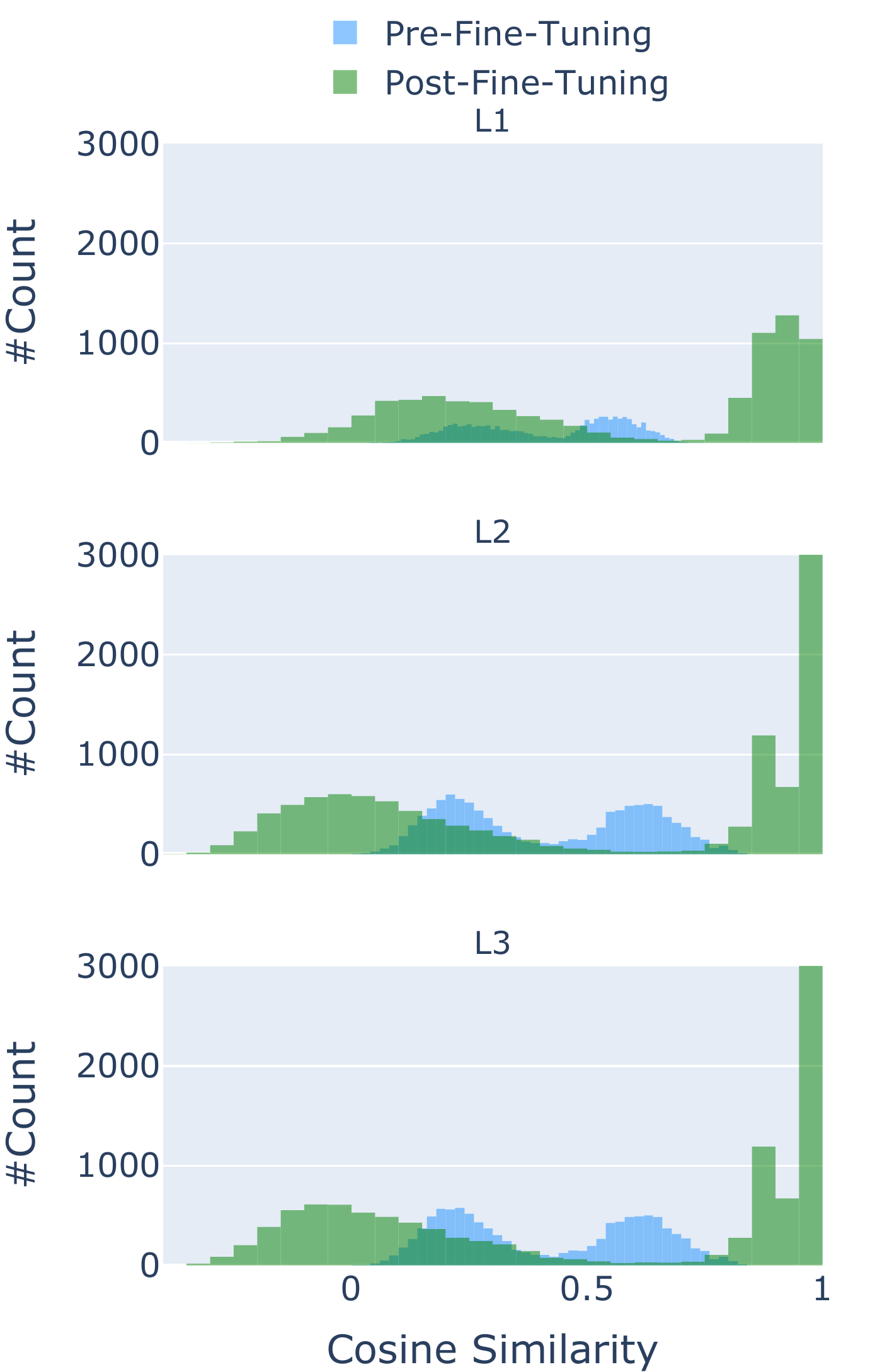}% Reduce the figure size so that it is slightly narrower than the column.
\caption{Histogram-based comparison of embedding space of a retriever pre- and post \newmodel\ fine-tuning across temporal task for positive and antagonist query pairs between Romanian and English.}
\label{fig:embedro}
\end{figure}

In this experiment, we endeavor to scrutinize the implications of fine-tuning a retriever model on its cross-lingual, time-sensitive semantic embedding space across languages and temporal tasks. The methodology employed utilizes a test set and S-BERT as foundational retriever to evaluate the aforementioned model. Precisely, our analysis is predicated on charting the temporal and semantic similarity among paired sets of cross-lingual queries, categorized into positive and antagonistic pairs. For the procurement of positive pairs, we systematically select queries in English and their corresponding translations in assorted low-resource languages. Conversely, the delineation of antagonistic pairs, herein referred to as 'antagonistic pairs,' is achieved by sampling an equal number of query pairs whose pre-fine-tuning semantic similarity scores do not exceed a threshold of $0.5$, presupposing these pairs exhibit a substantive dissimilarity.

\begin{figure}[t]
\centering
\includegraphics[width=0.4\textwidth]{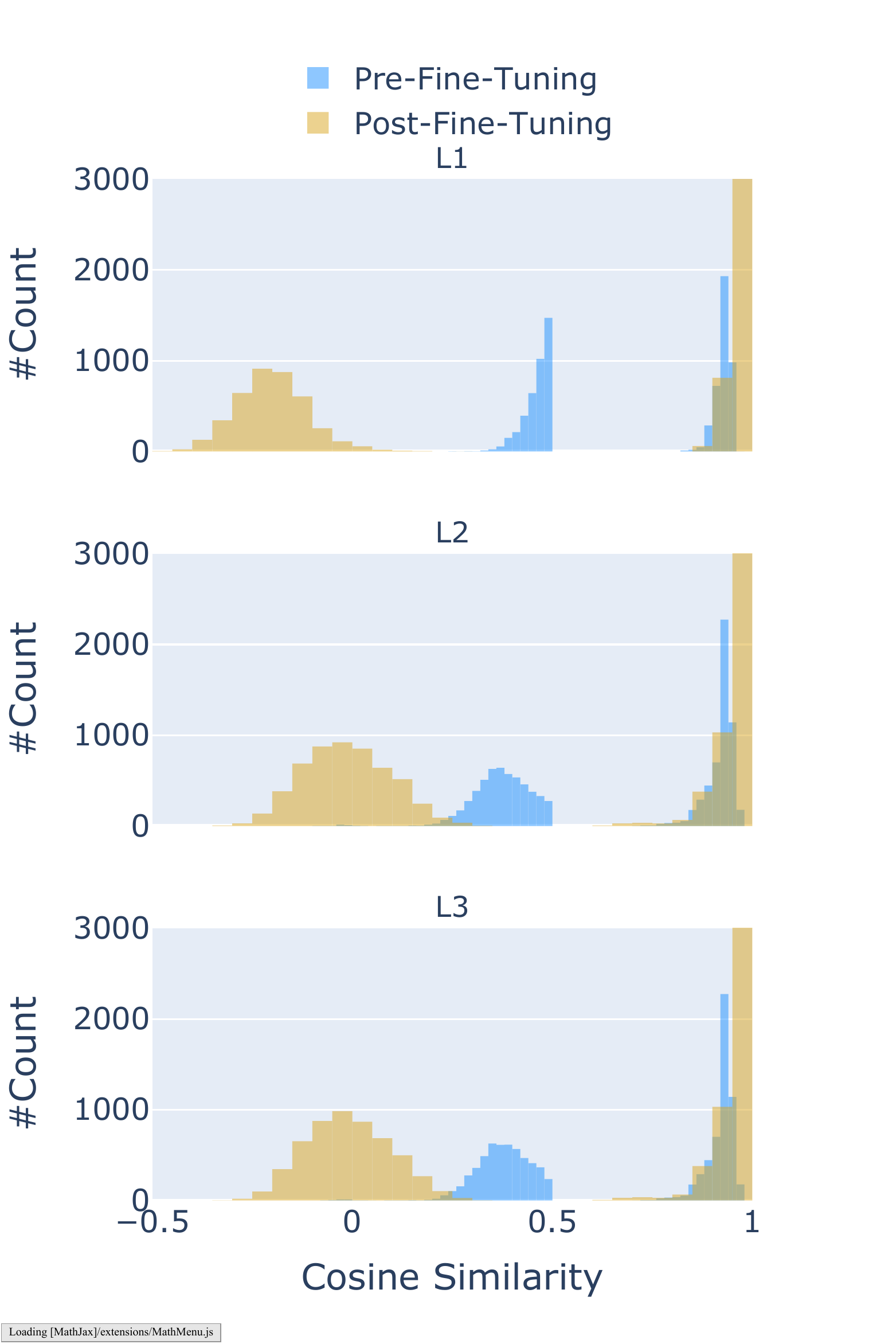}% Reduce the figure size so that it is slightly narrower than the column.
\caption{Histogram-based comparison of embedding space of a retriever pre- and post \newmodel\ fine-tuning across temporal task for positive and antagonist query pairs between German and English.}
\label{fig:embedge}
\end{figure}

\begin{figure}[t]
\centering
\includegraphics[width=0.4\textwidth]{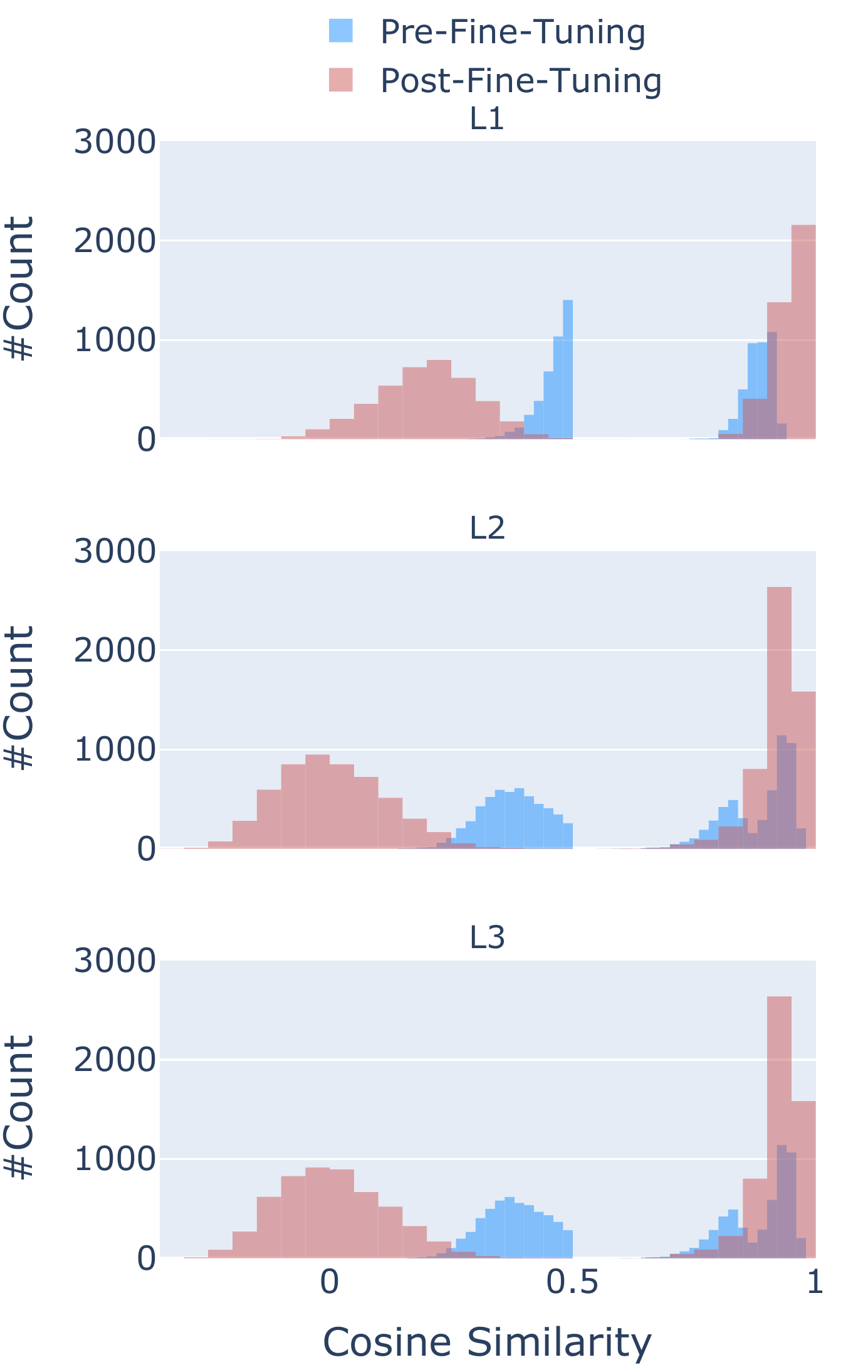}% Reduce the figure size so that it is slightly narrower than the column.
\caption{Histogram-based comparison of embedding space of a retriever pre- and post \newmodel\ fine-tuning across temporal task for positive and antagonist query pairs between French and English.}
\label{fig:embedfr}
\end{figure}

The empirical outcomes, as depicted in Figures referenced as Figure \ref{fig:embedro} (for Romanian), Figure \ref{fig:embedge} (for German), and Figure \ref{fig:embedfr} (for French), across varied temporal tasks, unequivocally demonstrate the augmented capacity of \newmodel, in heightening the semantic congruence for positively aligned query pairs while concurrently diminishing the semantic connection for antagonistic pairs. This enhancement in the model's embedding space unequivocally substantiates its improved performance in terms of F1 scores across the temporal tasks and low-resource languages.

\subsection{Hyperparameters}
In this study, the This study used the Huggingface\footnote{https://huggingface.co/} repository to stack various open-source large language models (LLMs). The development and fine-tuning of CLiTSSA were carried out using the PyTorch library, serving as the foundational framework. Minimal post-processing was applied to the outputs generated by the LLMs, entailing the removal of special characters and sequential indicators (e.g., "1)", "a)"), as well as the standardization of month names, particularly for Task L1, prior to the derivation of the final response. To fine-tune the \newmodel\ model, $1,000$ samples from the validation set and all samples from the training set were utilized to create a parallel corpus. To precisely capture the expected time-sensitive cross-lingual similarity distributions, the parameters h and w were heuristically determined to be $30$ and $10$, respectively. This configuration facilitated the generation of sufficient $40,000$ query pairs, each accompanied by their predicted similarity scores, across every task and language considered in this study. The dataset utilized for training the retrieval model was divided into two distinct portions: $10\%$ was reserved for validation purposes, whereas the remaining $90\%$ was earmarked for training. We use 'distiluse-base-multilingual-cased-v1' variant of multilingual Sentence-BERT across experiments for this study. %We contend that the primary role of parameters h and w is to ensure the generation of a sufficiently large sample size and capture the distribution necessary for the fine-tuning process; thus, they are not deemed critical for subsequent ablations. 
For the execution of these experiments, an NVIDIA A100\footnote{https://www.nvidia.com/en-in/data-center/a100/} GPU with 80GB memory was utilized.

\subsection{Extended Results: Robustness Across LLMs}
The results are presented in Table \ref{tab:llm_gen2}

\begin{table*}[!htb]
\centering
\small
%\resizebox{.95\columnwidth}{!}{
\begin{tabular}{l|l|c|c|c|c|c|c}
\toprule
     & &\multicolumn{3}{c}{\textbf{F1.}}&\multicolumn{3}{c}{\textbf{EM}} \\
    \cline{3-8}
    \textbf{Model} & \textbf{Method} & \textbf{L1} & \textbf{L2} & \textbf{L3}& \textbf{L1} & \textbf{L2} & \textbf{L3}\\
    \hline

\multirow{2}*{{Llama-3-8b}}& \xinsta$^\uparrow$ &46.62	&11.92	&17.74	&22.05&	4.55&	10.55\\
& \newmodel & \textbf{57.15}	&\textbf{15.23}&	\textbf{19.87}&	\textbf{32.57}&	\textbf{5.81}&	\textbf{11.22}\\
    \midrule
\multirow{2}*{{Mistral-v1}}& \xinsta$^\uparrow$ &38.26	&11.73&	\textbf{18.72}&8.75	&3.37	&12.20\\
& \newmodel & \textbf{46.45}&	\textbf{14.83}	&18.64&	\textbf{13.1}	&	\textbf{4.78}	&	\textbf{12.38}\\
    \midrule
\multirow{2}*{{Vicuna-7b-v1.5}}& \xinsta$^\uparrow$ &27.93	&9.54	&12.68&11.27	&1.92	&0.76\\
& \newmodel & \textbf{36.67}	&\textbf{12.04}	&\textbf{12.89}&	\textbf{18.8}	&	\textbf{2.05}	&0.76\\
    \midrule
\multirow{2}*{{Bloomz-7b1}}& \xinsta$^\uparrow$ &29.45	&3.79	&13.49&6.2	&	\textbf{0.81}	&9.64\\
& \newmodel &\textbf{40.74}	&\textbf{4.20}	&\textbf{13.57}&	\textbf{7.75}	&0.7	&	\textbf{9.89}\\
\midrule
&$\overline{\Delta}_{\newmodel-\uparrow}$ &\textcolor{blue}{9.68$\uparrow$}&\textcolor{blue}{2.33$\uparrow$}&\textcolor{blue}{$0.58\uparrow$}&\textcolor{blue}{5.98$\uparrow$}&	\textcolor{blue}{0.67$\uparrow$}&	\textcolor{blue}{0.27$\uparrow$}\\

\bottomrule
\end{tabular}
\caption{\newmodel\ performance across LLM's for temporal tasks using French test set. $\Bar{\Delta}$ represents the mean improvement in F1 score and EM (exact match)across LLM's for a temporal task.} 
\label{tab:llm_gen2}
\end{table*}

\subsection{Extended Results: Cross-Task CLiTSSA Performance}
The results are presented in Table \ref{tab:cross_task}

\begin{table*}[!htb]
\centering
%\resizebox{.95\columnwidth}{!}{
\begin{tabular}{l|c|c|c|c|c|c}
\toprule
 \textbf{} & \multicolumn{3}{c|}{\textbf{F1.}} & \multicolumn{3}{c}{\textbf{EM}}\\
 \cline{2-7}
    \textbf{Method} & \textbf{L1} & \textbf{L2} & \textbf{L3}& \textbf{L1} & \textbf{L2} & \textbf{L3} \\
    \midrule

{\xinsta}&\cellcolor{LightGray}{46.62}&\cellcolor{LightGray}{ 11.92}	&\cellcolor{LightGray}{17.74} &\cellcolor{LightGray}{22.05}	&\cellcolor{LightGray}{4.55}	&\cellcolor{LightGray}{10.55}\\
\newmodel\ [Fine-tuned with L1 Data]&\cellcolor{LightGray}{57.15}&	13.53	&20.23 &\cellcolor{LightGray}{32.57}	&5.39	&11.61\\
\newmodel\ [Fine-tuned with L2 Data]&46.49&	\cellcolor{LightGray}{15.23}	&20.15 &22.85	&\cellcolor{LightGray}{5.81}	&11.52\\
\newmodel\ [Fine-tuned with L3 Data]&45.15&	15.15	&\cellcolor{LightGray}{19.87} &21.47	&5.79	&\cellcolor{LightGray}{11.22}
\\

\bottomrule
\end{tabular}
\caption{Cross-task retriever performance across tasks with F1 scores and EM (exact match) metrics on the French test set against the X-InSTA baseline.
}
\label{tab:cross_task}
\end{table*}

\subsection{Extended Results: Cross-Lingual Versus Monolingual}
The results are presented in Table \ref{tab:mono}

\begin{table*}[t]
\centering
%\resizebox{.95\columnwidth}{!}{
\begin{tabular}{l|l|c|c|c|c|c|c}
\toprule
&  & \multicolumn{3}{c|}{\textbf{F1.}} & \multicolumn{3}{c}{\textbf{EM}} \\
\cline{3-8}

\textbf{Task}& \textbf{Settings} &\textbf{En$_m$} &\textbf{Fr$_m$}& \textbf{Fr$_c$}&\textbf{En$_m$} &\textbf{Fr$_m$}& \textbf{Fr$_c$}\\

\midrule

\multirow{4}*{L1} & Zero-Shot &  22.54	&11.83&	 \cellcolor{LightGray} &5.27 	&2.4& \cellcolor{LightGray}	\\	
 & \xicl  &76.85	&53.02&	33.60		& 59.95	
&6.57	&14.85	\\
 & \xinsta  & 83.94	&66.67&	46.62	&73.52	&30.07	
&22.05	\\
 & \newmodel  &\cellcolor{LightGray}	&\cellcolor{LightGray}&	\textbf{57.15}	&\cellcolor{LightGray}	&\cellcolor{LightGray}&\textbf{32.57}\\

%\cline{3-8}
&$\Delta_{Fr^{c,m}}$ &\cellcolor{LightGray} & \cellcolor{LightGray}& \textcolor{blue}{$\boldsymbol{-9.52\,(10.53\uparrow})$}&\cellcolor{LightGray}& \cellcolor{LightGray}&\textcolor{blue}{$\boldsymbol{+2.5\,(10.52\uparrow})$}\\
\midrule
&&&&&&&\\
\multirow{4}*{L2} & Zero-Shot  &8.84	&4.38	&\cellcolor{LightGray}	&3.07	&1.29	&\cellcolor{LightGray}	\\
 & \xicl &20.75	&14.67	&11.00	&7.61	&5.79 &4.05 	\\
 & \xinsta  &26.39	&20.20	&11.92	&12.56	&8.76	&3.57	\\
 & \newmodel  &\cellcolor{LightGray}	&\cellcolor{LightGray}&	\textbf{15.23}	&	\cellcolor{LightGray}&\cellcolor{LightGray}&\textbf{5.81}
\\

%\cline{3-8}
& $\Delta_{Fr^{c,m}}$&\cellcolor{LightGray} & \cellcolor{LightGray}& \textcolor{blue}{$\boldsymbol{-4.97\,(3.31\uparrow})$}&\cellcolor{LightGray}& \cellcolor{LightGray}&\textcolor{blue}{$\boldsymbol{-2.95\,(1.26\uparrow})$}\\
\midrule
&&&&&&&\\
\multirow{4}*{L3} & Zero-Shot &7.65	&16.99	&\cellcolor{LightGray}  	&1.94	&4.04	&\cellcolor{LightGray}	\\
 & \xicl &16.55	&18.46	&17.07		&4.90	&9.26	&7.28 \\
 & \xinsta&19.85	&21.33	&17.74		&11.63	&14.86	&10.55 	\\
 & \newmodel  &\cellcolor{LightGray}	&\cellcolor{LightGray}&	\textbf{19.87}	& \cellcolor{LightGray}	&\cellcolor{LightGray}
&\textbf{11.22}\\
&$\Delta_{Fr^{c,m}}$&\cellcolor{LightGray} & \cellcolor{LightGray}& \textcolor{blue}{$\boldsymbol{-1.46\,(2.13\uparrow})$}&\cellcolor{LightGray}& \cellcolor{LightGray}&\textcolor{blue}{$\boldsymbol{-3.64\,(0.67\uparrow})$}\\
\bottomrule
\end{tabular}
\caption{A comparative analysis of F1 scores and EM (exact-match) across temporal tasks in monolingual and cross-lingual scenarios utilizing LLaMA3-8B. Where \textit{En$_m$} and \textit{Fr$_m$} represent monolingual settings for English and French, respectively, while \textit{Fr$_c$} is French's cross-lingual setting. $\Delta_{Fr^{m,c}}$ representing the gap between \newmodel-based cross-lingual performance and \xinsta-based monolingual performance for French along with improvements over \xinsta-based cross-lingual setup within small brackets.}
\label{tab:mono}
\end{table*}

\end{document}